%% file: root.tex
\documentclass[letterpaper, 10 pt, conference]{ieeeconf}  

\IEEEoverridecommandlockouts                              

\usepackage{epsfig} 
\usepackage{caption}
\usepackage{subcaption}
\usepackage{booktabs}
\usepackage{multirow}
\usepackage{pgfplotstable}
\usepackage{times} 
\usepackage{amsmath} 
\usepackage{amssymb}  
\usepackage{hyperref}
\usepackage{graphicx}
\usepackage[outdir=./]{epstopdf}
\usepackage{xcolor}
\usepackage{xspace}
\usepackage{flushend}
\input{shortcuts.tex}

\title{\LARGE \bf
Learning Cascaded Detection Tasks with \\
Weakly-Supervised Domain Adaptation
}

\author{Niklas Hanselmann$^{1,2}$, Nick Schneider$^{1}$, Benedikt Ortelt$^{3}$ and Andreas Geiger$^{2,4}$
\thanks{*This publication was created as part of the research project "KI Delta Learning" (project number: 19A19013A) funded by the Federal Ministry for Economic Affairs and Energy (BMWi) on the basis of a decision by the German Bundestag.}
\thanks{$^{1}$Mercedes-Benz AG, R\&D, Stuttgart, Germany}%
\thanks{$^{2}$University of T\"ubingen, T\"ubingen, Germany}%
\thanks{$^{3}$Robert Bosch GmbH, Stuttgart, Germany}
\thanks{$^{4}$Max Planck Institute for Intelligent Systems, T\"ubingen, Germany}
\thanks{Primary contact: {\tt\small niklas.hanselmann@daimler.com}}
}

\begin{document}

\maketitle
\thispagestyle{empty}
\pagestyle{empty}

\input{abstract.tex}

\input{sections/introduction.tex}


\input{sections/related_work.tex}

\input{sections/method.tex}

\input{sections/experiments.tex}

\input{sections/conclusion.tex}



\bibliographystyle{IEEEtran}
\bibliography{bibliography_short,bibliography_custom,bibliography}

\end{document}

%% file: shortcuts.tex
\makeatletter
\DeclareRobustCommand\onedot{\futurelet\@let@token\@onedot}
\def\@onedot{\ifx\@let@token.\else.\null\fi\xspace}

\def\eg{\emph{e.g}\onedot} 
\def\ie{\emph{i.e}\onedot}

\def\etal{\emph{et al}\onedot}
\makeatother

\newcommand{\figref}[1]{Fig.~\ref{#1}}
\newcommand{\tabref}[1]{Tab.~\ref{#1}}
\newcommand{\secref}[1]{Sec.~\ref{#1}}

\newcommand{\ba}{\mathbf{a}}
\newcommand{\bb}{\mathbf{b}}
\newcommand{\bc}{\mathbf{c}}

\newcommand{\bx}{\mathbf{x}}

\newcommand{\bz}{\mathbf{z}}

\newcommand{\nR}{\mathbb{R}}

\newcommand{\cA}{\mathcal{A}}
\newcommand{\cB}{\mathcal{B}}
\newcommand{\cC}{\mathcal{C}}
\newcommand{\cD}{\mathcal{D}}

\newcommand{\cL}{\mathcal{L}}

\newcommand{\cS}{\mathcal{S}}
\newcommand{\cT}{\mathcal{T}}

\newcommand{\cX}{\mathcal{X}}

\newcommand{\cZ}{\mathcal{Z}}

\DeclareMathOperator*{\argmax}{argmax~}
\DeclareMathOperator*{\argmin}{argmin~}

\makeatletter
\DeclareRobustCommand\onedot{\futurelet\@let@token\@onedot}
\def\@onedot{\ifx\@let@token.\else.\null\fi\xspace}
\def\eg{e.g\onedot} 
\def\ie{i.e\onedot}

\def\etal{et~al\onedot}

\makeatother

\newcommand{\boldparagraph}[1]{\vspace{0.2cm}\noindent{\bf #1:}}

\definecolor{darkgreen}{rgb}{0,0.7,0}



%% file: abstract.tex
\begin{abstract}
In order to handle the challenges of autonomous driving, deep learning has proven to be crucial in tackling increasingly complex tasks, such as 3D detection or instance segmentation. State-of-the-art approaches for image-based detection tasks tackle this complexity by operating in a cascaded fashion: they first extract a 2D bounding box based on which additional attributes, e.g. instance masks, are inferred. While these methods perform well, a key challenge remains the lack of accurate and cheap annotations for the growing variety of tasks. Synthetic data presents a promising solution but, despite the effort in domain adaptation research, the gap between synthetic and real data remains an open problem. In this work, we propose a weakly supervised domain adaptation setting which exploits the structure of cascaded detection tasks. In particular, we learn to infer the attributes solely from the source domain while leveraging 2D bounding boxes as weak labels in both domains to explain the domain shift. We further encourage domain-invariant features through class-wise feature alignment using ground-truth class information, which is not available in the unsupervised setting. As our experiments demonstrate, the approach is competitive with fully supervised settings while outperforming unsupervised adaptation approaches by a large margin.
\end{abstract}

%% file: sections/introduction.tex
\section{Introduction}
Modern deep learning-based vision systems have made remarkable strides in recent years, enabled
in no small part by leveraging large amounts of labeled training data.
Unfortunately, acquiring annotated data is often expensive, time consuming and, for some tasks,
also inaccurate or nearly impossible.
To alleviate this problem, synthetic data obtained from game-engines or purpose-built
simulation environments might be an alternative, promising a large amount of training data with
highly accurate, inexpensive annotations. However, despite ever-improving photorealism, current
state-of-the-art synthetic datasets still fail to resemble actual real world sensor data.
Subsequently, machine learning models which are trained on a synthetic dataset typically perform poorly if
applied to real world data.
In order to bridge this domain gap, several works on domain adaptation
have been proposed. Most commonly, they assume an \textit{unsupervised} regime, in which we have access to
fully-labeled data in a source domain and only unlabeled data in a target domain.
This setting is particularly difficult and while the community continues to make progress in
tackling the open problems in unsupervised domain adaptation, the gap in
performance when compared to target domain supervision remains significant, limiting
the utility of these models for practical applications, such as autonomous driving. Is there an alternative which
allows for accurate results without the need of expensive labeling in the target domain?

In this paper, we propose a setting which exploits the decomposability of complex detection tasks into a cascaded structure. This concept
is central to many image-based detection approaches where objects are first detected and, based on the detection,
additional attributes such as instance masks \cite{He2017ICCV} or 2D/3D poses \cite{AlpGueler2018CVPR} are inferred.
Likewise, labeling can now also be decomposed
into 2D bounding box labels, which are inexpensive to annotate and for which annotated datasets already exist in many domains, and additional, more complex attribute labels,
which are often expensive or difficult to obtain with sufficient accuracy. Motivated by this, we
investigate the use of weak supervision in the form of 2D bounding box annotations in the target domain in conjunction with full supervision
in the source domain. Our key hypothesis is that a large part of the domain shift might already be explained in the
detection stage, while the following cascaded stage transfers more readily given accurate detections. To test this hypothesis, we consider two common
cascaded detection tasks in this paper: Instance segmentation and monocular 3D detection.

\begin{figure}[t]
	\centering
	\includegraphics[width=0.48\textwidth]{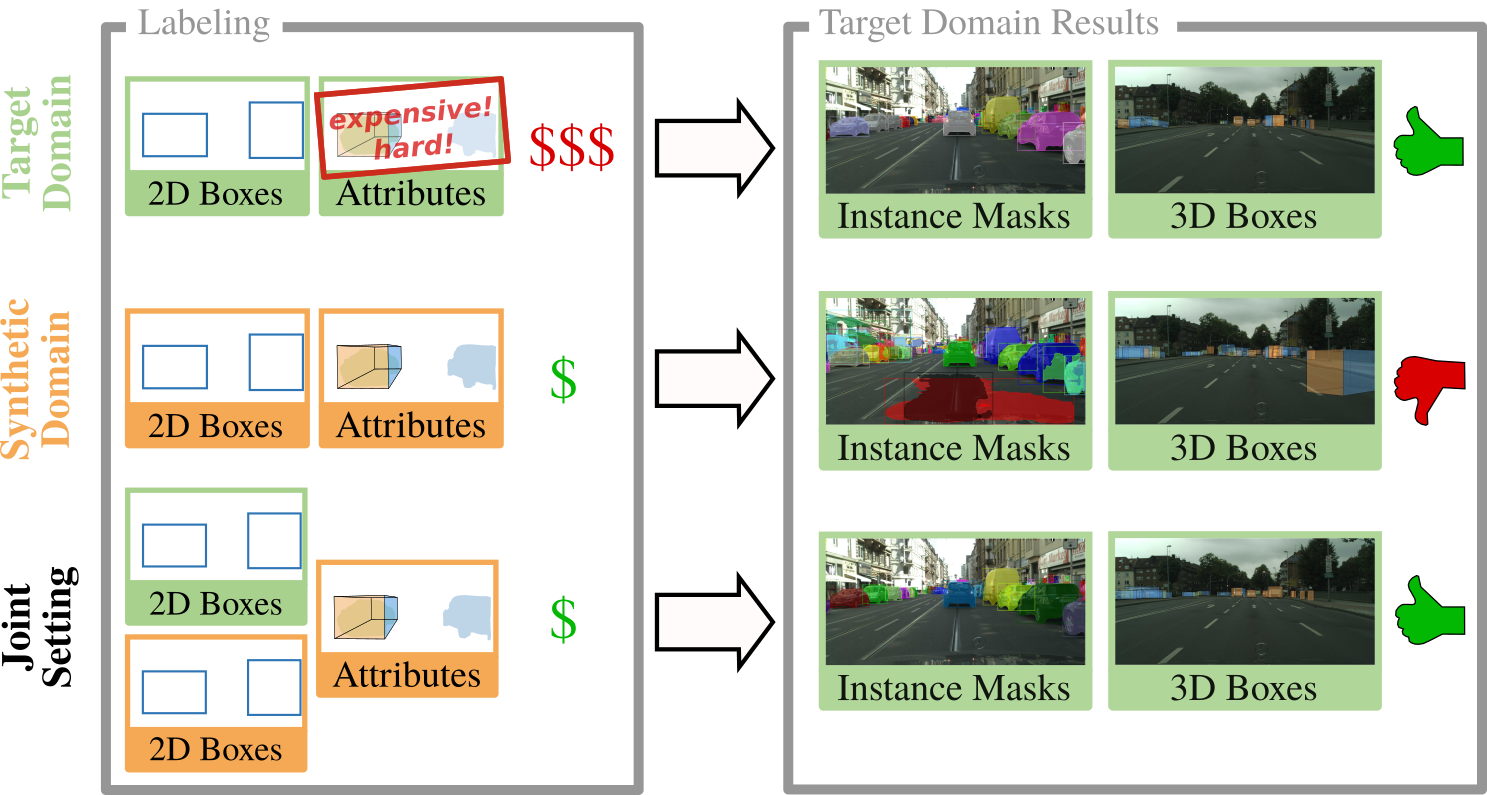}
	\caption{\textbf{Proposed Setting.}
        We propose a weakly-supervised domain-adaptation setting that enables learning cascaded detection tasks at a reduced annotation effort while still achieving competitive performance by utilizing weak 2D bounding box labels in both domains.
	}
	\label{fig:teaser}
\end{figure}

Although we find that a strategy of jointly training on both domains with the available weak
supervision can already work surprisingly well, there is no guarantee
that the learned feature representation will always be useful for the cascaded stage in the target domain.
As evidenced by our experiments in \secref{sec:experiments}, joint training can in fact lead to
domain-specific features, rendering the cascaded stage less transferable.
Borrowing from the unsupervised domain adaptation literature, we address this problem by encouraging
domain-invariant representations. A typical approach is to align marginal feature distributions across domains.
However, this does not guarantee a low target domain error for the specific task and can even be detrimental
if the label distributions differ across domains \cite{Zhao2019ICML}.
Recent work has addressed this problem by considering class information during feature
alignment \cite{Long2018NIPS, Luo2019CVPR}, aiming at invariance of the conditional feature distributions
given the corresponding labels. However, doing so in the absence of
ground-truth target domain labels is challenging due to error accumulation \cite{Jiang2020ICML}.
Our setting allows us to sidestep this issue as it enables us to accurately condition the
feature distribution alignment on the ground-truth
class at instance-level.

In summary, we make the following contributions:
\begin{itemize}
    \item
    We propose and systematically analyze a weakly supervised domain adaptation setting for cascaded detection tasks,
    where 2D bounding box annotations are available in both domains.
    \item
    We show that in this setting, models adapted using existing techniques can now be competitive with models that are fully
    supervised in the target domain at a significantly reduced annotation effort.
    Furthermore, they outperform unsupervised adaptation approaches by a large margin, justifying the additional
    supervision.
    \item
    We analyze the role of weak supervision in bridging the domain shift both in isolation
    and in conjunction with feature distribution alignment.
\end{itemize}

With this work we hope to both inspire and provide a baseline for future research on domain adaptation
leveraging weak labels. Supplementary material is available at: \url{https://lasnik.github.io/wsda/}.

%% file: sections/related_work.tex
\section{Related Work}
The domain adaptation problem has received great interest in recent years, especially in the
unsupervised setting, with several studies in the context of diverse tasks
such as classification \cite{Ganin2015ICML, Long2018NIPS}, semantic segmentation \cite{Bousmalis2017CVPR,Hoffman2018ICML,Coors2019THREEDV}
and more recently object detection \cite{Chen2018CVPRd, Saito2019CVPR,Xu2020CVPR}. In the following,
we review the most related work on unsupervised domain adaptation in general and in the context of
object detection specifically, as well as cross-domain learning
with weak supervision.

\boldparagraph{Unsupervised Domain Adaptation}
Underpinned by the theory proposed in \cite{Ben-David2010}, the majority of approaches in the
unsupervised regime tackle domain adaptation by attempting to mitigate the discrepancy between data
distributions across domains. Building on the progress in generative modeling and
image-to-image translation \cite{Isola2017CVPR, Zhu2017ICCVa}, one line of work seeks to achieve this directly in
image space \cite{Hoffman2016ARXIVcustom, Bousmalis2017CVPR,Hoffman2018ICML}, aiming at
distributional alignment at \textit{pixel-level}. A popular
alternative has been to instead focus on alignment at \textit{feature-level}.
Methods in this category typically attempt to match the marginal
distributions of source and target domain features. This is done either explicitly by
minimizing a divergence measure \cite{Tzeng2014ARXIVcustom, Sun2016ARXIV} or
implicitly through domain-adversarial training \cite{Ganin2015ICML, Long2018NIPS}.
Orthogonal to these approaches, another line of work leverages unlabeled
auxiliary tasks and utilizes them in a self-training scheme alongside the main task
\cite{Hoffman2016ARXIVcustom, Zhang2017ICCV, Lian2019ICCV}.
Many of these works introduced concepts that are now central to the domain adaptation literature,
but focus mostly on image classification and semantic segmentation, while we are interested in
detection-based tasks.

\boldparagraph{Cross-Domain Object Detection}
In \cite{Chen2018CVPRd}, Chen \textit{\etal} follow the domain-adversarial training paradigm in a
first study of unsupervised domain-adversarial adaptation in the context of object detection: They
propose to align features at both image- and instance-level by means of
domain-adversarial training. Motivated by the hypothesis that transferability
of features might decrease towards deeper layers, Saito \textit{\etal} \cite{Saito2019CVPR}
employ strong and weak alignment strategies at image-level for shallow and deep
features, respectively. Several works follow a similar notion by proposing different
hierarchical alignment schemes \cite{He2019ICCV, Chen2020CVPR, Zheng2020CVPR, Hsu2020ECCV}, while
others focus on mining descriptive region- \cite{Zhu2019CVPRa} or instance-level \cite{Xu2020CVPR}
features for alignment. To ensure discriminativity of the learned representation, recent work
\cite{Zheng2020CVPR, Xu2020CVPR} proposes aligning category prototype embeddings by
relying on the detection model's class estimates in the target domain. In \cite{Wang2019CVPRb},
a few-shot adaptive cross-domain detection setting is studied, where a small number of annotated
target domain images are available, for which the per-class feature distributions are aligned with
ground-truth class labels. In a slight departure from the prevalent paradigm of
learning domain-invariant representations, \cite{Khodabandeh2019ICCV, Kim2019ICCV}
view the cross-domain detection problem as one of robust learning from
self-generated noisy pseudo-labels.
Although this body of work marks extensive
progress, current unsupervised domain adaptation methods still fail to close the
domain gap - often by a large margin - rendering them of limited use for practical
applications.

\boldparagraph{Cross-Domain Learning with Weak Supervision}
While the unsupervised regime has been extensively studied, domain adaptation
with access to weak annotations from an auxiliary task in the target domain has received
comparatively less attention.
Previous works utilize depth
images as weak supervision to aid in transferring a semantic segmentation model. As
the relationship between depth and semantics is not trivial,
these works focus mainly on strategies to transfer knowledge both across domains as well as between the
auxiliary task and the task of interest, either by learning direct mapping functions \cite{Ramirez2019ICCV}
or through knowledge distillation \cite{Zhou2020ECCV}. Although an intriguing direction,
like unsupervised approaches, these early works are unsuccessful in closing the domain gap.
In \cite{WANG2019TIP}, Wang \textit{\etal} also consider the task of semantic segmentation, but leverage
2D bounding boxes as weak supervision in both domains. They consider a multitask detection and semantic segmentation
architecture, where both tasks share a single backbone network and apply class-agnostic image-level
and class-wise instance-level adversarial feature alignment. Here again, the exact relationship between
detection and semantic segmentation is not immediately obvious, yielding results comparable to
unsupervised domain adaptation methods. In contrast to these works, we specifically consider a cascaded setting
where the auxiliary task is \emph{subsumed} by the task of interest and its potential benefit is more
readily apparent.
Similarly to ours, another line of work follows this same notion, for example by leveraging
attribute annotations for fine-grained recognition \cite{Gebru2017ICCV},
image-level class annotations for object detection \cite{Inoue2018CVPR} or 2D pose annotations for
3D human pose estimation \cite{Zhou2017ICCV}. These works consider a specific task
and rely on explicit constraints induced by its relationship to the auxiliary task, such as consistency between
attribute-level and fine-grained predictions, the ability to generate pseudo-labels from weak annotations
or geometric correspondences between poses in 2D and 3D.
In contrast, we are interested
in leveraging 2D bounding box annotations for an entire class of related problems, namely multiple
cascaded detection tasks, without focusing on explicit, task-specific regularization.

%% file: sections/method.tex
\section{Method}
\label{sec:method}
In this section, we present our weakly-supervised domain adaptation approach. First,
we formally describe the learning problem resulting from our novel setting in \secref{subsec:learning_problem}.
In order to solve this problem,
we use the cascaded framework described in \secref{subsec:cascaded_model}, which allows to
leverage weak annotations in the target domain. Finally, in
\secref{subsec:joint_training} and \ref{subsec:feature_alignment}, we describe strategies to optimize
the cascaded detection model in our setting with the aim of robust and accurate target domain
performance.

\begin{figure*}[ht]
	\centering
	\includegraphics[width=0.95\textwidth]{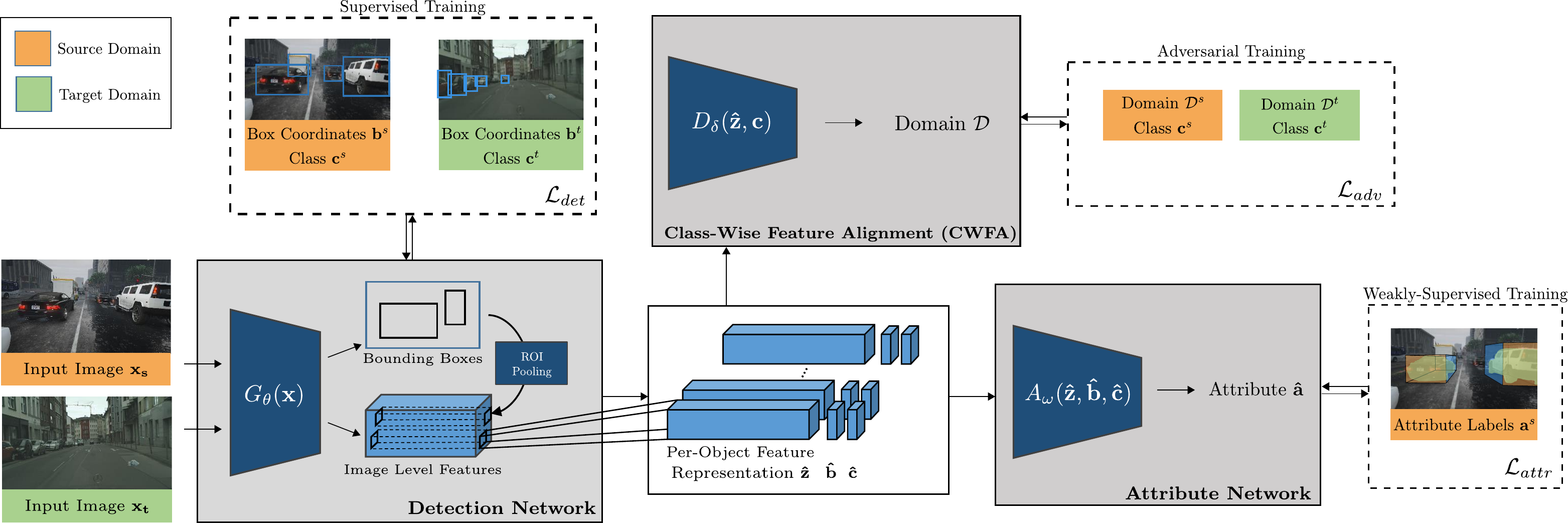}
    \caption{\textbf{Overview of the presented approach.}
    The goal of our approach is to detect objects in a target domain image by describing them in terms of their location
    $\bb$ and class $\bc$ as well as additional, more complex attributes $\ba$, such as precise shape
    or 3D information. We consider a weakly supervised domain adaptation setting, where labels for
    the attributes are only available in the source domain (\eg synthetic data), while
    easier-to-obtain 2D bounding boxes are available in both domains. We utilize a cascaded detection
    framework where first a detection network $G_{\theta}$, trained with supervision on both domains,
    extracts $\hat{\bb}$ and $\hat{\bc}$ as well as per-object
    feature representations $\hat{\bz}$. Based on the detection network's outputs the attribute
    network $A_{\omega}$, supervised on the source domain only, then infers $\hat{\ba}$. To encourage representations that are meaningful in both
    domains, the distributions over features $\hat{\bz}$ are aligned via a class-conditional domain classifer $D_{\delta}$.
    }
	\label{fig:overview}
\end{figure*}

\subsection{Learning Problem}
\label{subsec:learning_problem}
Image-based object detection aims to extract the image location, typically in the form of
2D bounding box coordinates $\bb \in \cB \subset \nR^{k \times 4}$
, and semantic
class information $\bc \in \cC = \{c_0, c_1,..., c_{L-1}\}^k$ of all $k$ objects in the image, where L is the
number of classes under consideration.
There exist various extensions to the task that additionally estimate more complex attributes
$\ba$ for each object, such as precise shape or 3D information.
However, these additional attributes are often expensive or difficult to label and may therefore not be available in the target domain $\cT$.
A natural solution is to use training data from another domain $\cS$, which contains the missing labels.

Let $\cD^s$ and $\cD^t$ denote the source and target domain datasets respectively, with
$
    \cD^s = \left\{
        \left(
            \bx_i^s, \bb_i^s, \bc_i^s, \ba_i^s
        \right)
    \right\}_{i=1}^{N_s}
$, where $\bx_i^s$ is the input image and $N_s$ is the number of samples in the dataset.
The notation follows analogously for the target domain
$
    \cD^t = \left\{
        \left(
            \bx_i^t, \bb_i^t, \bc_i^t
        \right)
    \right\}_{i=1}^{N_t}
$
, where we have access to weak supervision in the form of 2D bounding box coordinates $\bb$ and class information $\bc$,
but no supervision for the additional attributes.
We assume that both domains share a common label space, but follow different data
distributions.
Given $\cD^s$ and $\cD^t$, our goal is then to learn a cascaded detection model with
source domain supervision on the attributes $\ba$
that will transfer well to the target domain, while utilizing the supervision on $\bb$ and $\bc$
that is available in both domains.

While the approach is applicable to any cascaded detection task, we focus on the problems
of instance segmentation and monocular 3D detection as examples in this work.
Consequently, the additional attributes
$\ba$ represent pixel-wise instance masks and 3D shape and pose for instance segmentation and 3D detection, respectively.

\subsection{Cascaded Detection Model}
\label{subsec:cascaded_model}
For the cascaded detection framework, we follow the state of the art \cite{He2017ICCV, Simonelli2019ICCV}
and choose a multi-stage architecture,
a design choice that also follows naturally from our setting, as it explicitly encodes the
decomposition of the cascaded detection task. We hence have two main components, (1) a base
network responsible for 2D detection and (2) an attribute network that builds on the detection
network's predictions to estimate additional attributes for each detection. In the following we
use the hat operator to denote model predictions.

\boldparagraph{Detection network}
The detection network $G_{\theta}$ is a convolutional neural network with parameters $\theta$, that maps the input
image $\bx \in \cX$ to estimates of the per-object 2D bounding boxes $\hat{\bb} \in \cB$ and semantic class labels $\hat{\bc} \in \cC$:
\begin{equation}
    G_{\theta}(\bx) : \cX \to \cB \times \cC \times \cZ
\end{equation}
It consists of a backbone, which extracts an image-level feature representation from the
input, and a detection module, that predicts $\hat{\bb}$ and $\hat{\bc}$ from that representation.
In addition to 2D
detections, this network also provides a per-object feature representation $\hat{\bz} \in \cZ$
for the attribute network by applying ROIAlign \cite{He2017ICCV}
on the image-level representation using the estimated bounding box coordinates $\hat{\bb}$.

\boldparagraph{Attribute network}
The attribute network $A_{\omega}$ is a neural network with parameters $\omega$ that, given detections
described by $\hat{\bb}$ and $\hat{\bc}$, estimates the additional attributes $\hat{\ba} \in \cA$ from their object-level
feature representation $\hat{\bz}$:
\begin{equation}
    A_{\omega}(\hat{\bz}, \hat{\bb}, \hat{\bc}) : \cB \times \cC \times \cZ \to \cA
\end{equation}
The specific instantiation of both $\ba$ and consequently $A_{\omega}$ depends on the cascaded
detection task under consideration.

\subsection{Joint Training with Weak Supervision}
\label{subsec:joint_training}
Given this framework, how do we best utilize $\cD^s$ and $\cD^t$ to learn the optimal model parameters $\theta^*$ and $\omega^*$, such
that the resulting model will be robust and accurate in the target domain?
Let us begin by considering the unsupervised domain adaptation setting, where we have to rely
solely on source domain supervision, resulting in the following combined loss function for both the
detection and the attribute network:
\begin{equation}
    \begin{split}
    \cL_{source} =
        \frac{1}{N_s}\sum_{i=1}^{N_s} &\cL_{det}(G_{\theta}(\bx_i^s), \bb_i^s, \bc_i^s) + \\
        &\cL_{att}\left(A_{\omega}\left(G_{\theta}\left(\bx_i^s\right)\right), \ba_i^s\right)
    \end{split}
\end{equation}
The exact formulation of $\cL_{det}$ and $\cL_{att}$ depends on the specific instantiations of
$G_{\theta}$ and $A_{\omega}$, which are described in \secref{subsec:experiment_setup}.
In our setting, we also assume access to target domain supervision for the detection network:
\begin{equation}
    \cL_{target} =
        \frac{1}{N_t}\sum_{i=1}^{N_t} \cL_{det}\left(G_{\theta}\left(\bx_i^t\right), \bb_i^t, \bc_i^t\right)
\end{equation}
We denote the sum of both loss terms as $\cL_{joint}$, yielding the following combined objective:
\begin{equation}
    \left(\theta^*, \omega^*\right) = \argmin_{\theta, \omega} \cL_{joint}
\end{equation}
where the detection network is trained jointly with supervision from both domains. A key hypothesis in
this work is that such a joint training strategy can potentially aid
the transferability of the cascaded detection model across domains in two main aspects:
Firstly,
since any prediction of the attributes $\hat{\ba}$ implicitly assumes a valid and accurate 2D detection,
the performance of the detection network is crucial. Training it jointly on both domains
ensures that this assumption is less likely to be broken, decoupling the adaptation efforts for the
attribute network from the 2D detection task. Secondly, the detection network might learn shared
representations across domains when trained jointly, rendering any attribute network
making predictions based on those representations more readily transferable.

\subsection{Class-Wise Feature Alignment}
\label{subsec:feature_alignment}
\label{cfda}
Although we find that this joint training strategy works well in some scenarios,
it may not always be optimal.
Firstly, its success depends highly on the degree to which the attribute
network reuses the feature representations that emerge by training the detection network jointly.
If both networks build on distinct, task-specific features in the backbone
such a strategy cannot hope to improve cross domain performance beyond increased accuracy of the
detection network. Secondly, there is no guarantee that joint training
always results in a shared representation across domains. Should the representations indeed be
domain-specific, the attribute network might not yield the desired results in the target domain.

To counteract this, we follow the same paradigm of regularizing the learned
representation to be domain-invariant that is central to many state-of-the-art unsupervised
cross-domain detection methods.
Since in our setting ground-truth information on the object coordinates and semantic
class is available in the target domain, we can perform accurate object-level class-wise
feature alignment (CWFA) without error accumulation. To this end, we utilize a class-conditional domain
classifier $D_{\delta}$, which tries to separate the features according to their
domain given their ground-truth class. At the same time the detection network $G_{\theta}$ is optimized
to output features that maximally confuse the domain classifier in an adversarial fashion.
More concretely, $D_{\delta}$ maps the object-level features $\hat{\bz}$ to a probability of belonging
to the target domain for each class and then outputs the probability associated with the current
ground-truth class $\bc$:
\begin{equation}
    D_{\delta}(\hat{\bz}, \bc) : \cZ \times \cC \to \left[0, 1\right]
\end{equation}
This approach of considering class information is similar to previous work in the context of
few-shot cross-domain detection
\cite{Wang2019CVPRb} and is particularly effective in our proposed setting, as we have the
ground-truth class available for every example. Following \cite{Saito2019CVPR},
we use a focal loss \cite{Lin2017ICCVcustom} with weighting parameter
$\gamma$ to optimize the domain classifier:
\begin{equation}
\begin{split}
    &\cL_{adv} = \\
    &-\frac{1}{N_s}\sum_{i=1}^{N_s} 
    \left(1 - D_{\delta}\left(G_{\theta}\left(\bx_i^s\right), \bc_i^s\right)\right)^\gamma 
    \log{D_{\delta}\left(G_{\theta}\left(\bx_i^s\right), \bc_i^s\right)} \\ 
    &-\frac{1}{N_t}\sum_{i=1}^{N_t} 
    D_{\delta}\left(G_{\theta}\left(\bx_i^t\right), \bc_i^t\right)^\gamma 
    \log{\left(1 - D_{\delta}\left(G_{\theta}\left(\bx_i^t\right), \bc_i^t\right)\right)}\\ 
\end{split}
\end{equation}
We thus obtain the following overall objective:
\begin{equation}
    \left(\theta^*, \omega^*, \delta^*\right) = \argmin_{\theta, \omega} \argmax_{\delta} \cL_{joint} - \lambda_{adv} \cL_{adv}
\end{equation}
where $G_{\theta}$ and $A_{\omega}$ are optimized to minimize both the 2D- and attribute detection
losses according to the joint training strategy while simultaneously maximally confusing the domain
classifier $D_{\delta}$, which is optimized for low domain classification error, and $\lambda_{adv}$ is a trade-off
parameter. As in previous works \cite{Chen2018CVPRd, Saito2019CVPR, Zheng2020CVPR}, the
adversarial min-max game is realized through the use of a
Gradient Reversal Layer (GRL) \cite{Ganin2015ICML} at the input of the domain classifier, which
flips the sign of the gradients and scales them by a weighting parameter $\lambda_{grl}$ during the backward pass.

%% file: sections/experiments.tex
\section{Experiments}
\label{sec:experiments}
\input{figures/instance_seg_comparison.tex}
In the following, we present our experimental results on various synthetic
and real datasets. To this end, we first describe our experimental setup in
\secref{subsec:experiment_setup} and subsequently compare our method to several baseline approaches
on the tasks of instance segmentation and monocular
3D detection in \secref{subsec:experiments:comparison_state_of_the_art}. Finally, we provide
an ablation study in \secref{subsec:experiments:ablation}.

\subsection{Experiment Setup}
\label{subsec:experiment_setup}
\boldparagraph{Datasets}
We perform experiments on four datasets:
(1) \textbf{\textit{Cityscapes (CS)}} \cite{Cordts2016CVPR} is a scene understanding benchmark containing
$5$k diverse real-world urban driving scenes.
(2) \textbf{\textit{Foggy Cityscapes (FCS)}} \cite{Sakaridis2018ECCV} is an extension to CS, that
adds synthetic fog to the original scenes. As a result, both have exactly the same label distribution.
(3) \textbf{\textit{Synscapes (SYN)}} \cite{Wrennige2018} is a collection of $25$k
photorealistic synthetic urban traffic scenes from a custom rendering enginge.
(4) \textbf{\textit{VIPER (VIP)}} \cite{Richter2017} contains over $100$k synthetic images
obtained from the video game \textit{Grand Theft Auto V}. \newline
With these datasets we build
three adaptation scenarios: SYN $\to$ CS, VIP $\to$ CS and
CS $\to$ FCS. For each scenario, we create a common label space between both domains. Where
possible, we correct differences in labeling policies during preprocessing, otherwise removing incompatible
classes from consideration.

\boldparagraph{Evaluation Protocol}
We report the performance on the validation split of the target domain
dataset using the official evaluation protocols \cite{Cordts2016CVPR, Gaehlert2020ARXIV}.
We use the mean average precision (mAP) and mean detection score (mDS) as metrics for
instance segmentation and monocular 3D detection, respectively. We run each experiment three times
using different random seeds and report the performance of the best run at the last training iteration.

\boldparagraph{Baselines \& Comparison}
As a lower bound we train a model on the source domain only and apply
it on the target domain during testing. As an upper bound, we train an oracle model with full
target domain supervision. In addition, we also consider the unsupervised
cross-domain detection approaches of Chen \etal \cite{Chen2018CVPRd} and Saito \etal \cite{Saito2019CVPR}.
We re-implemented their work in order to extend it to our datasets and setting. We compare these baselines
against models adapted in our proposed setting using both just the weakly-supervised joint training strategy
in isolation (\textit{WSJT}) and in conjunction with class-wise feature alignment (\textit{WSJT + CWFA}).
On the task of monocular 3D detection, we additionaly report results of the latter
configuration with \textit{class-agnostic} feature alignment (\textit{WSJT + CAFA}), for which we
use the instance-level alignment module in \cite{Chen2018CVPRd}, to evaluate the benefit of considering class information.
To the best of our knowledge, there is no prior work on leveraging 2D bounding box annotations for domain adaptation in the
context of instance segmentation and monocular 3D detection, thus we provide an additional ablation
study of the proposed weakly supervised setting in \secref{subsec:experiments:ablation}.

\input{figures/box3d_comparison.tex}
\input{results/instance_segmentation/all_datasets_all_methods.tex}
\input{results/bounding_box_3d/all_datasets_all_methods.tex}
\input{results/ablation_studies/box2d_gt_ablation_instance_segmentation.tex}
\boldparagraph{Implementation and Training Details}
As the multi-stage cascaded detection framework we use a standard implementation\footnote{https://github.com/facebookresearch/detectron2}
of Mask-RCNN \cite{He2017ICCV} with a Resnet-50 \cite{He2015ICCV}
backbone in conjunction with a Feature Pyramid Network (FPN) \cite{Lin2017CVPRacustom}. For monocular
3D detection, we replace the mask head with a reimplementation of the 3D bounding box regressor
proposed in \cite{Simonelli2019ICCV}, where we omit the confidence rescoring branch. This regressor works on amodal 2D bounding boxes which, in contrast to other datasets \cite{Caesar2019ARXIVcustom, Qi2019CVPR},
are currently not available in Cityscapes. We therefore compute them as projections of the 3D bounding boxes onto the image plane.
For the class-conditional domain classifier we reuse the instance-level architecture of Chen \etal \cite{Chen2018CVPRd}
and extend the last fully-connected layer to output $L$ per-class domain probabilities.
We adapt the methods of Chen \etal \cite{Chen2018CVPRd} and Saito \etal \cite{Saito2019CVPR} to the FPN configuration for a fair comparison by applying the corresponding
domain classifier to each FPN-level. Experiments using dedicated domain classifiers
for each FPN-level showed no benefit over this strategy. For instance segmentation, we use the
same loss functions for $\cL_{det}$ and $\cL_{att}$ as well as the same optimizer, learning rate,
training schedule and batch size as in \cite{He2017ICCV}. For monocular 3D
detection, we again adopt the hyperparameters and loss functios for $\cL_{det}$ and $\cL_{att}$
from \cite{Simonelli2019ICCV}. However, we use a reduced batch size of $8$ due to memory constraints and a
longer schedule, training for a total of $120$k
iterations and decaying the learning rate by a factor of $0.1$ at $72$k and
$90$k iterations.

\subsection{Main Results}
\label{subsec:experiments:comparison_state_of_the_art}
\input{figures/failure_cases.tex}

\boldparagraph{Instance Segmentation}
\label{subsec:experiments:instance}
For instance segmentation we consider the adaptation scenarios SYN $\to$ CS, VIP $\to$ CS and CS $\to$ FCS.
As shown in \tabref{tab:instance:all},
models trained in our setting outperform the unsupervised baselines by a large margin.
This is also mirrored in the
qualitative results shown in \figref{fig:instseg_comparison}, where our approach yields more defined and
accurate segmentations. While our approach is competitive with the fully supervised oracle for
SYN $\to$ CS and CS $\to$ FCS, there remains a performance gap of 5.8 percentage points for
VIP $\to$ CS, despite substantial gains over the baselines. Here,
the largest discrepancy (13.1 percentage points) is observed in the car class, which we attribute to an incompatibility in labeling
policies: contrary to Cityscapes, vehicle segmentation masks in VIPER do not include windows, leading
to false negatives in the target domain, which is illustrated in \figref{fig:failure_cases}.
Furthermore, we find that class-wise
feature alignment does not improve the results for instance segmentation. This suggests that
joint training of the base network does indeed promote a domain-invariant representation,
which the instance segmentation module in turn reuses.
Moreover, we observe that the unsupervised baseline methods only achieve a performance gain over the
source only baseline in the CS $\to$ FCS scenario, where the distribution of classes and scene layouts is identical.
In both remaining scenarios they do not significantly improve the performance. We
attribute this to negative transfer effects resulting from
the class-agnostic alignment of features for domains with distinct distributions over scene layouts
and relative frequencies of classes.
\label{subsec:experiments:3d}

\boldparagraph{Monocular 3D detection}
For monocular 3D detection we consider the adaptation scenarios SYN $\to$ CS and CS $\to$ FCS.
From \tabref{tab:detection_3d:all} we see that,
compared to instance segmentation, monocular 3D detection is significantly more difficult:
in this case, the simple joint training strategy still outperforms the source only and unsupervised
baselines, but is unsuccessful in fully closing the gap towards the oracle. This suggests that for
this task, the attribute network does not rely as strongly on the features learned via joint
training. We observe that here, additionally performing feature alignment does improve performance, which
supports this hypothesis. A possible explanation is that instance segmentation inherently synergizes more strongly with the detection task,
as both estimate object representations in image space (\ie 2D bounding boxes and instance masks),
while monocular 3D detection estimates object representations in 3D space (\ie 3D bounding boxes).
Furthermore, class-wise feature alignment does indeed outperform its class-agnostic counterpart,
verifying the benefit of utilizing the ground-truth class information available in our setting.
As indicated by the qualitative results in
\figref{fig:box3d}, our approach yields more accurate pose estimations compared to
the baselines, but fails for scenes which have a layout that differs from those found in the source
domain, \eg one-way streets (\figref{fig:failure_cases}).

\subsection{Ablation Study}
\label{subsec:experiments:ablation}
\boldparagraph{Performance over amount of weak target domain annotations}
In \figref{fig:map_over_time}, we compare the mAP over annotation time when using
our setting compared to full supervision with annotated instance masks in the target domain.
In internal experiments we observe that annotating the full task takes six times longer than annotating 2D bounding boxes
for both 3D detection and instance segmentation, which for the latter matches previous reports \cite{Bellver2019CVPRWORK}.
The results show that up to a critical amount of labeled
images, our approach is more efficient compared to fully supervised methods.
\input{results/ablation_studies/map_over_time.tex}

\boldparagraph{Ground-truth 2D boxes at test-time}
One of the main questions in this work is if the network is able to learn transferable features for
the cascaded detection module or if any observed performance gain is simply due to improved 2D detections.
In \tabref{tab:ablation:2d_gt}, we therefore analyze the performance of the instance segmentation and 3D box estimation
modules when decoupled from the performance of the 2D detector by using ground-truth 2D bounding boxes as input
during testing.
While for instance segmentation joint training already yields good results, the 3D bounding box
detector strongly benefits from the class-wise feature alignment. This supports our hypothesis that
for instance segmentation, joint training already learns transferable features, while for
monocular 3D detection further improvements can be made by explicitly aligning the feature distributions.

%% file: figures/instance_seg_comparison.tex
\begin{figure*}[tbp]
	\centering
	\begin{subfigure}[t]{0.22\textwidth}
		\centering
		\includegraphics[trim=0 0 0 100, clip=true, width=\linewidth]{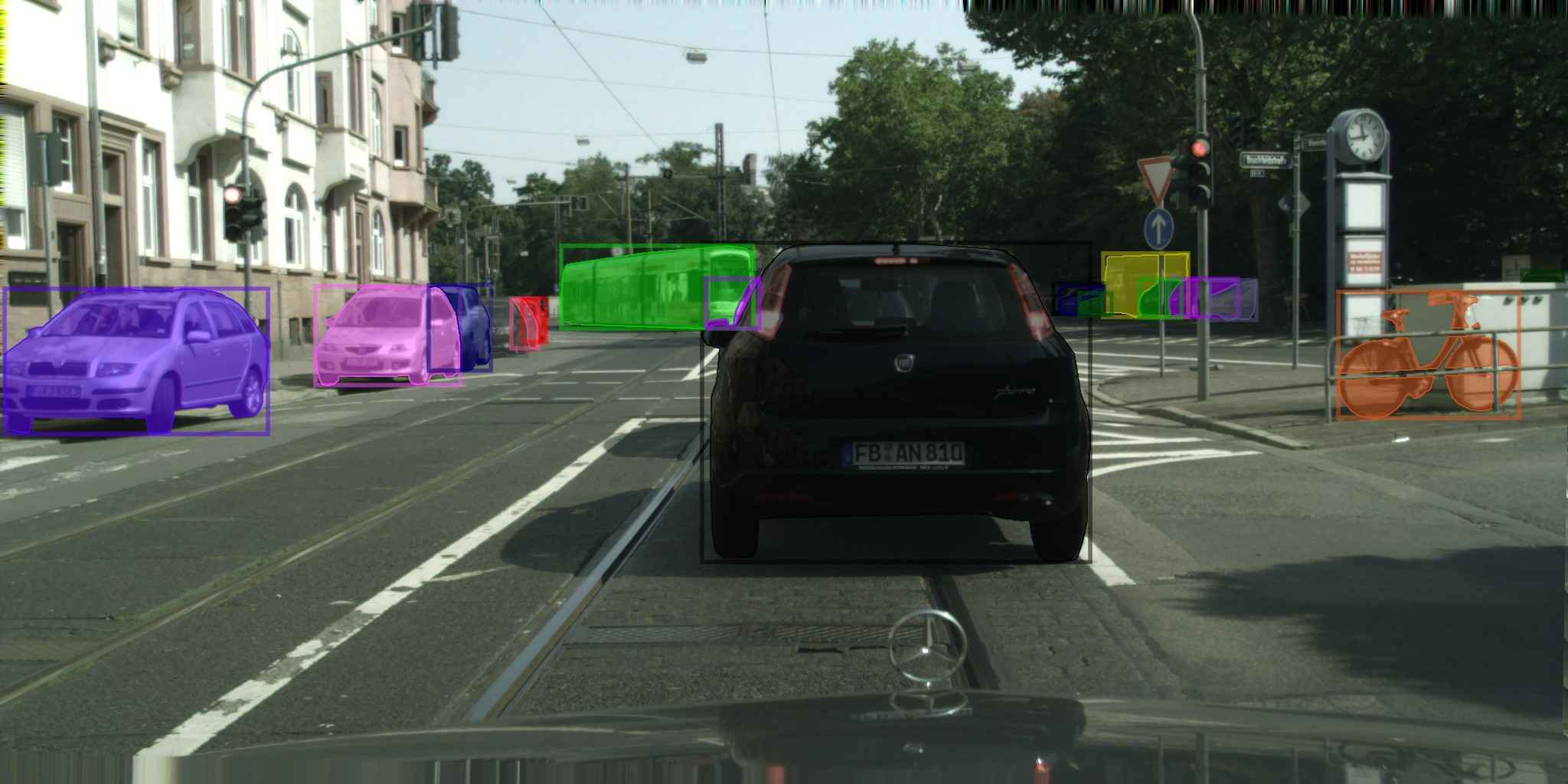}
	\end{subfigure}
	\begin{subfigure}[t]{0.22\textwidth}
		\centering
		\includegraphics[trim=0 0 0 100, clip=true, width=\linewidth]{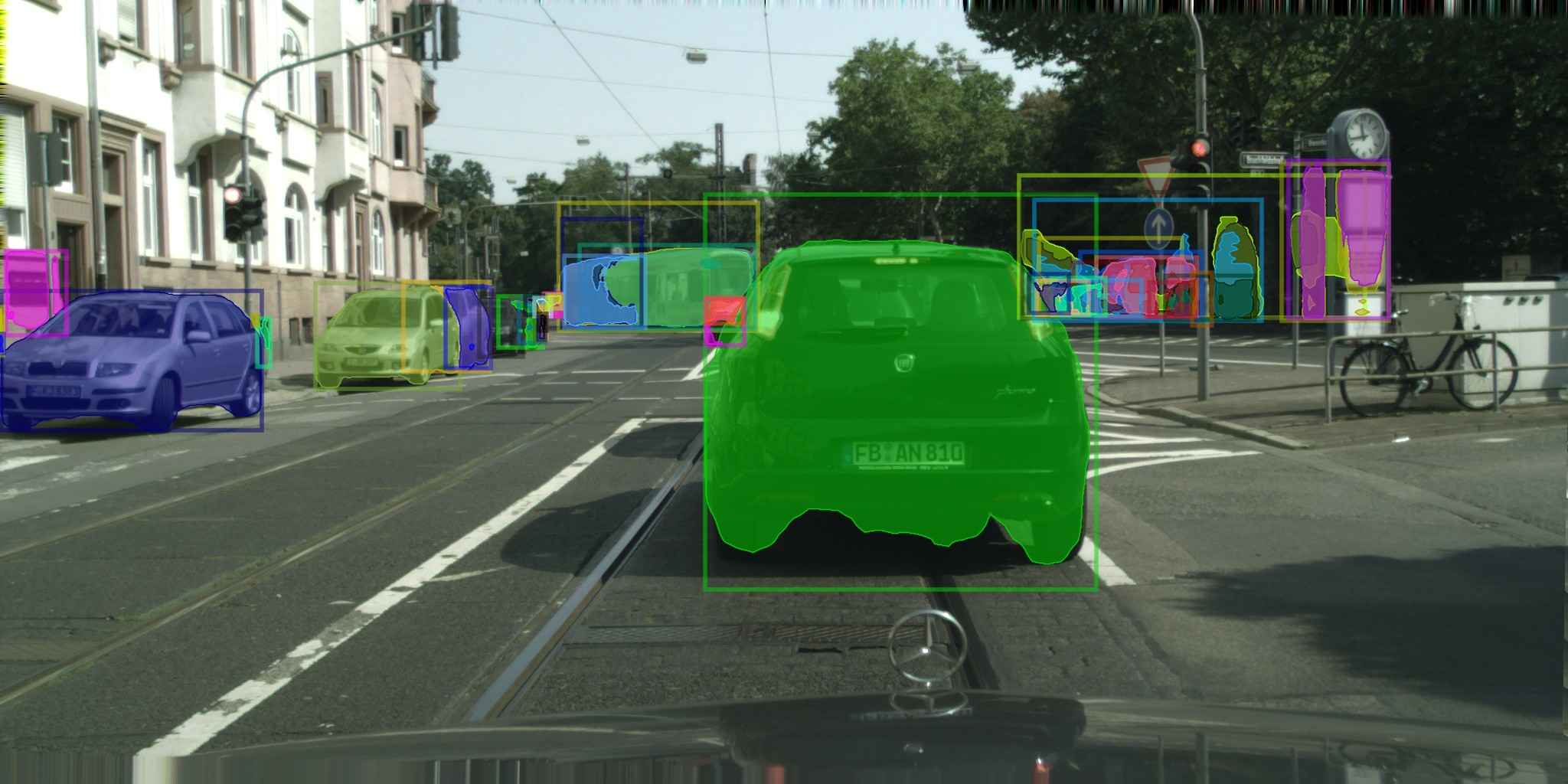}
	\end{subfigure}
	\begin{subfigure}[t]{0.22\textwidth}
		\centering
		\includegraphics[trim=0 0 0 100, clip=true, width=\linewidth]{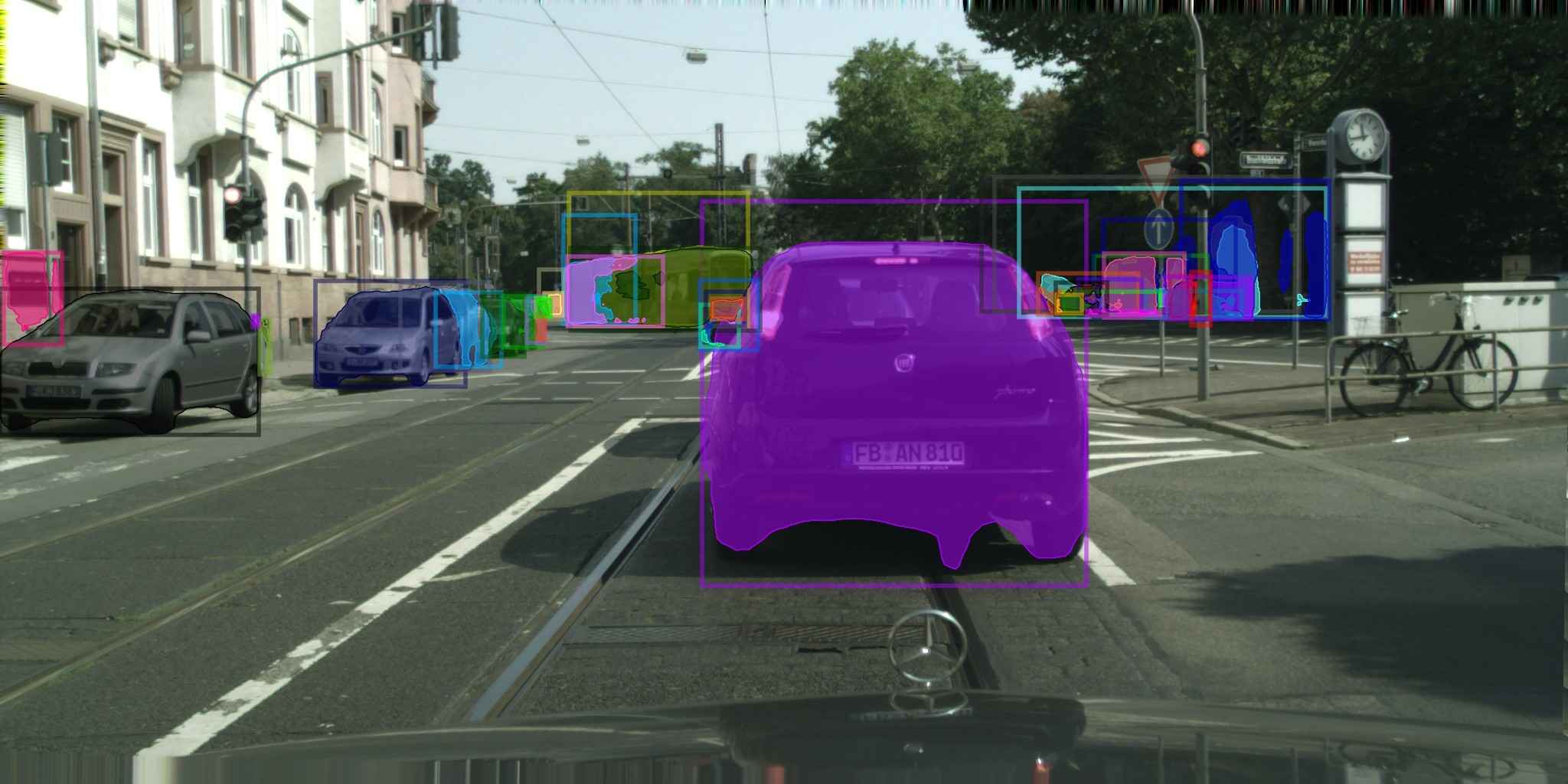}
	\end{subfigure}
	\begin{subfigure}[t]{0.22\textwidth}
		\centering
		\includegraphics[trim=0 0 0 100, clip=true, width=\linewidth]{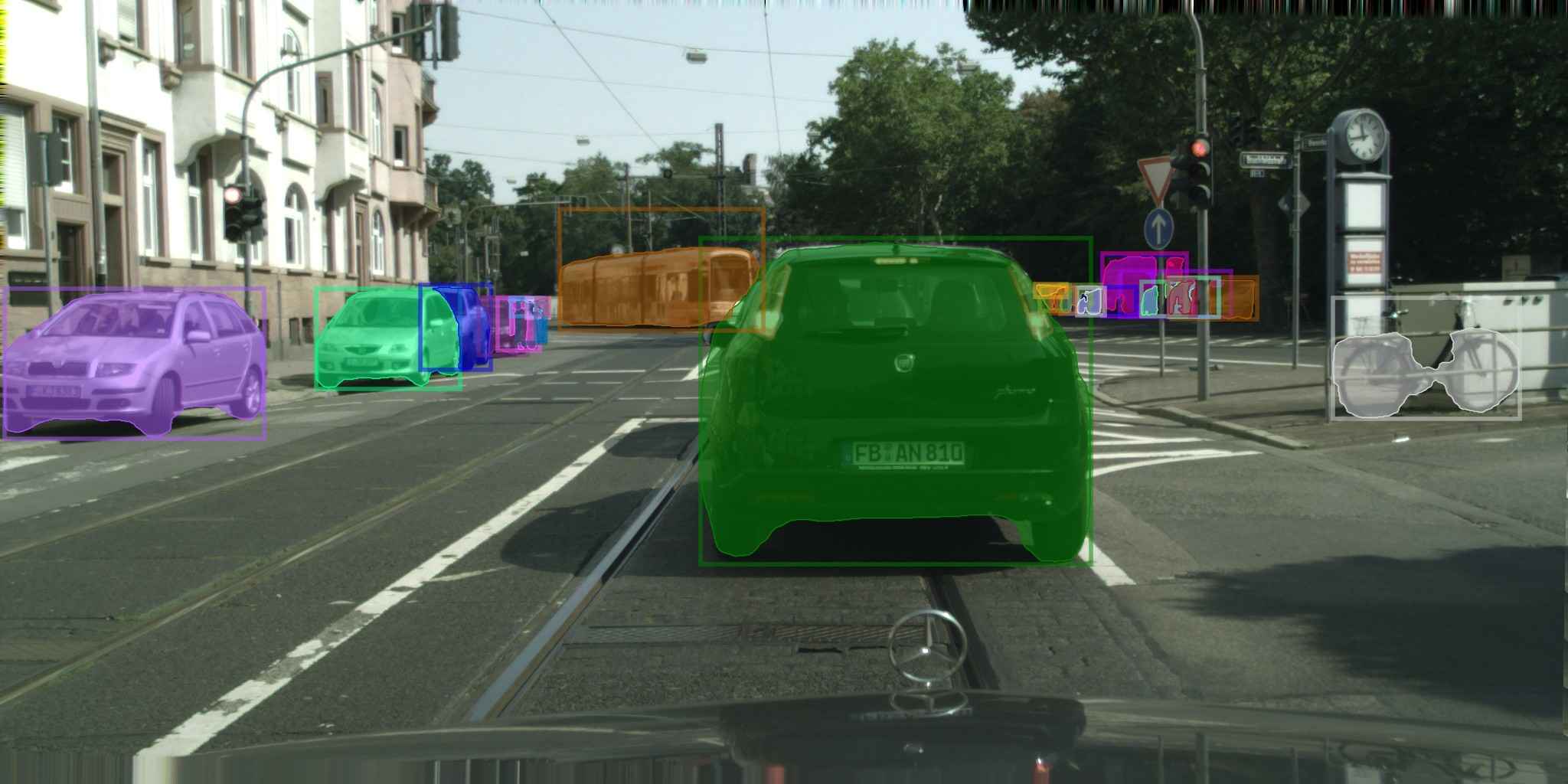}
	\end{subfigure}
	\begin{subfigure}[t]{0.22\textwidth}
		\centering
		\includegraphics[trim=0 0 0 100, clip=true, width=\linewidth]{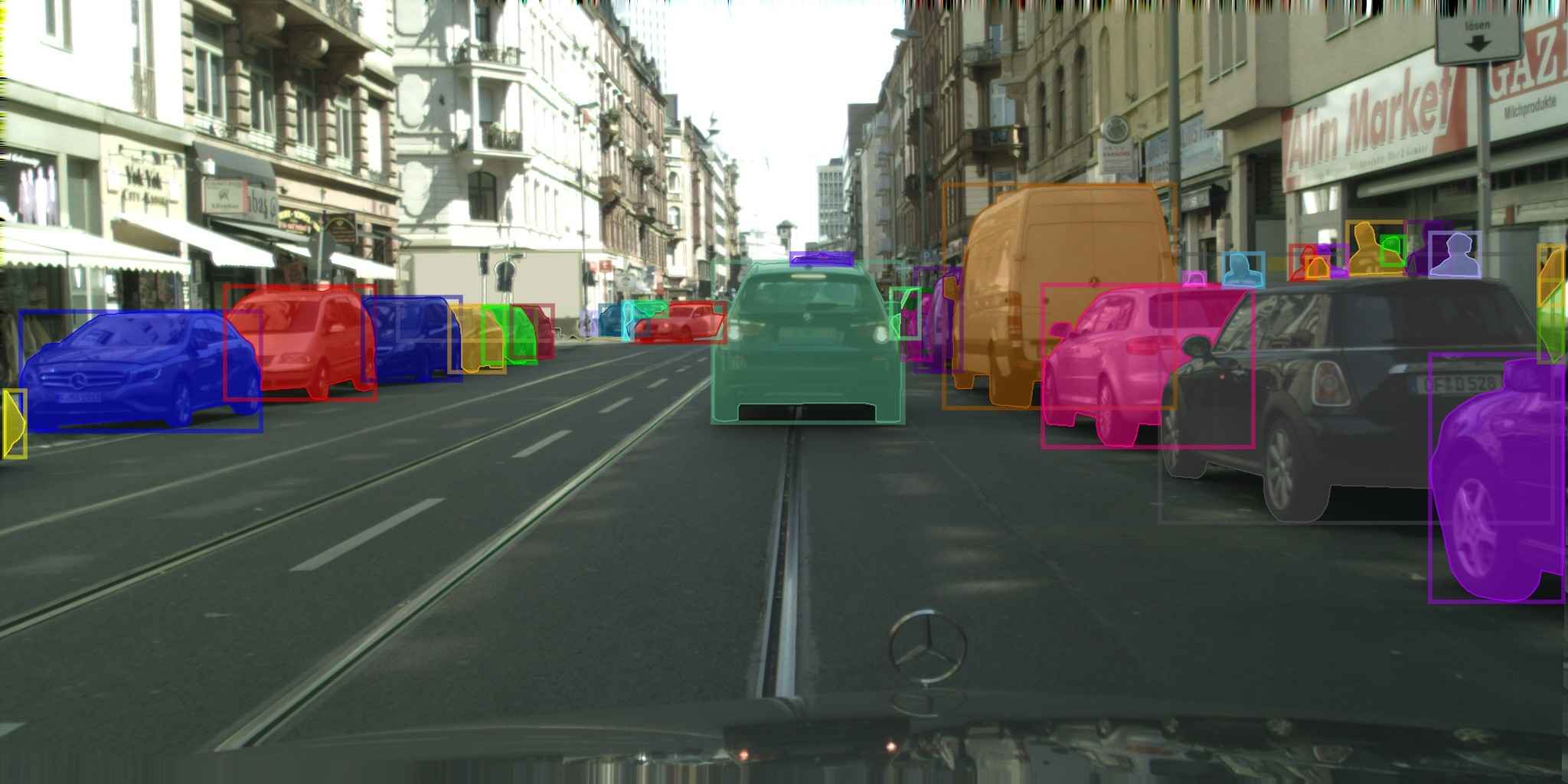}
	\end{subfigure}
	\begin{subfigure}[t]{0.22\textwidth}
		\centering
		\includegraphics[trim=0 0 0 100, clip=true, width=\linewidth]{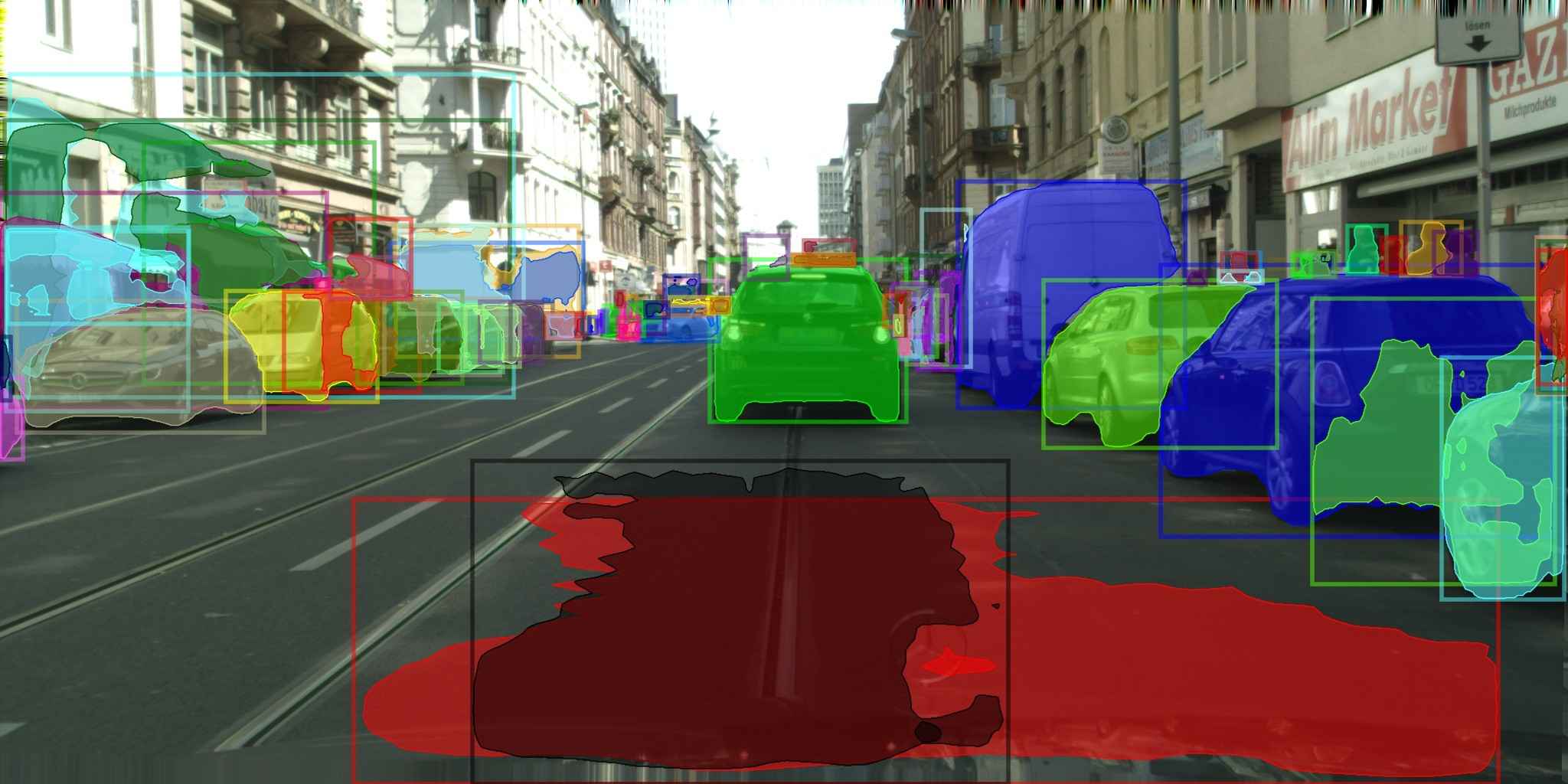}
	\end{subfigure}
	\begin{subfigure}[t]{0.22\textwidth}
		\centering
		\includegraphics[trim=0 0 0 100, clip=true, width=\linewidth]{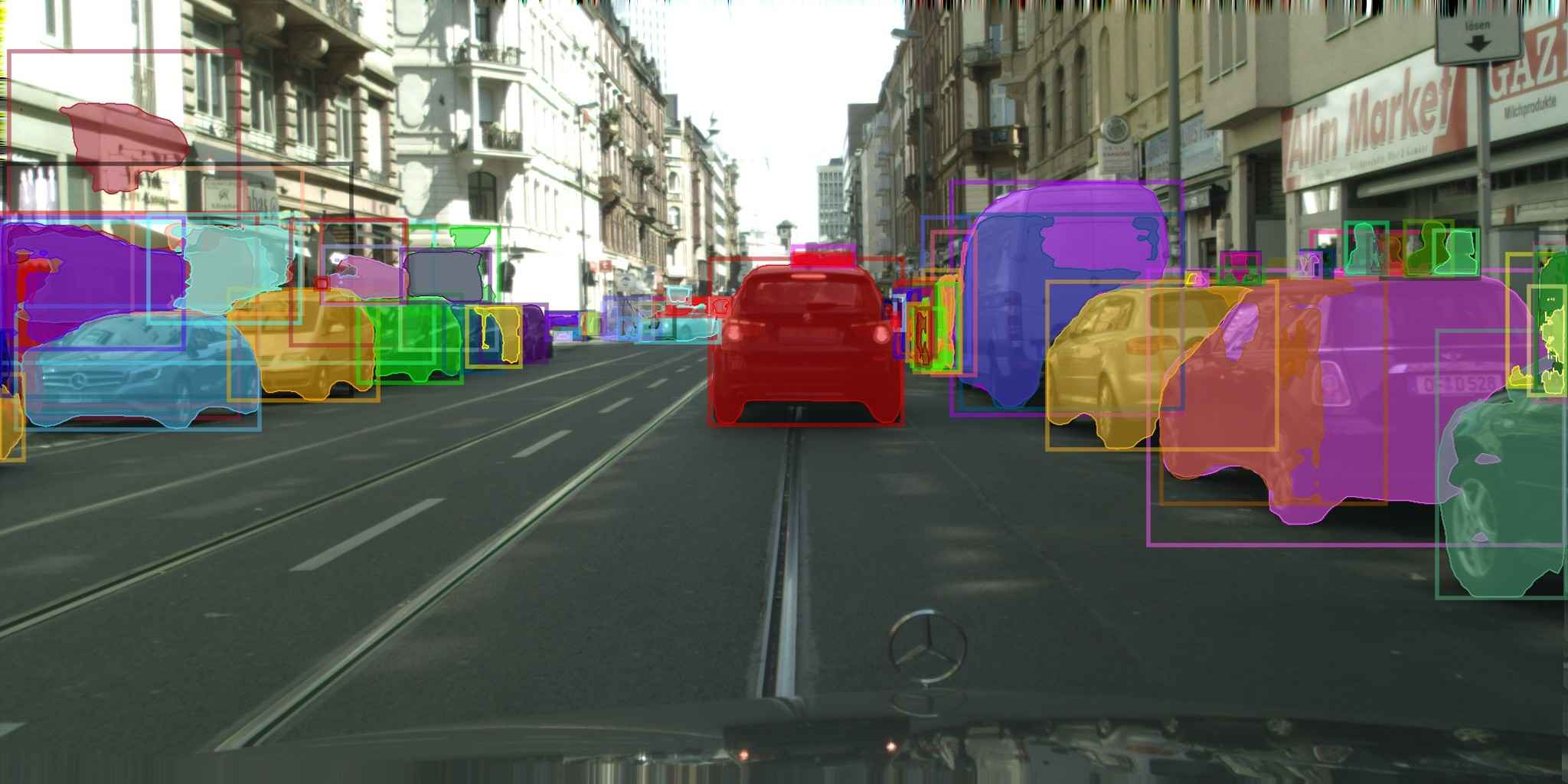}
	\end{subfigure}
	\begin{subfigure}[t]{0.22\textwidth}
		\centering
		\includegraphics[trim=0 0 0 100, clip=true, width=\linewidth]{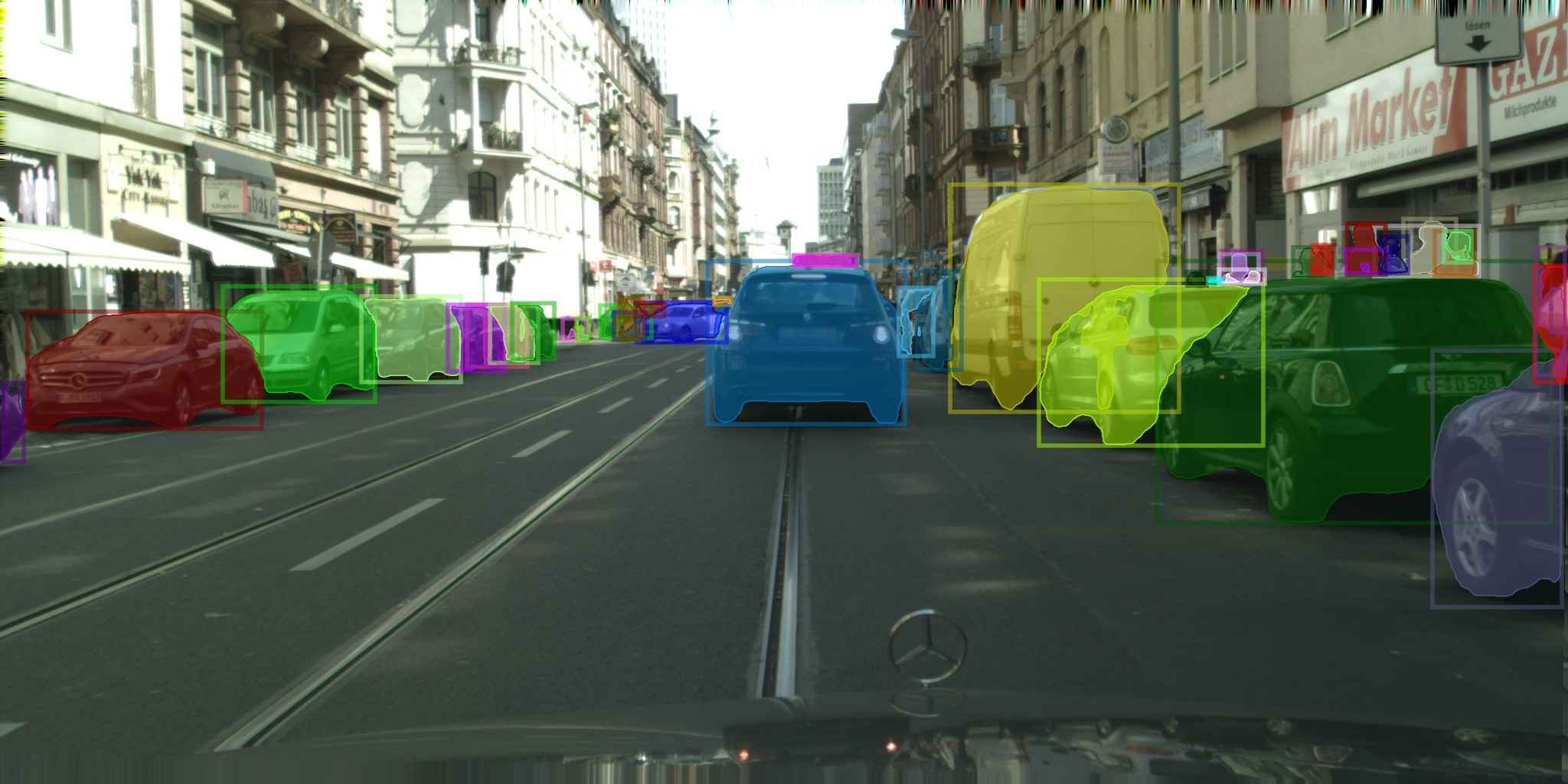}
	\end{subfigure}
	\begin{subfigure}[t]{0.22\textwidth}
		\centering
		\includegraphics[trim=0 0 0 100, clip=true, width=\linewidth]{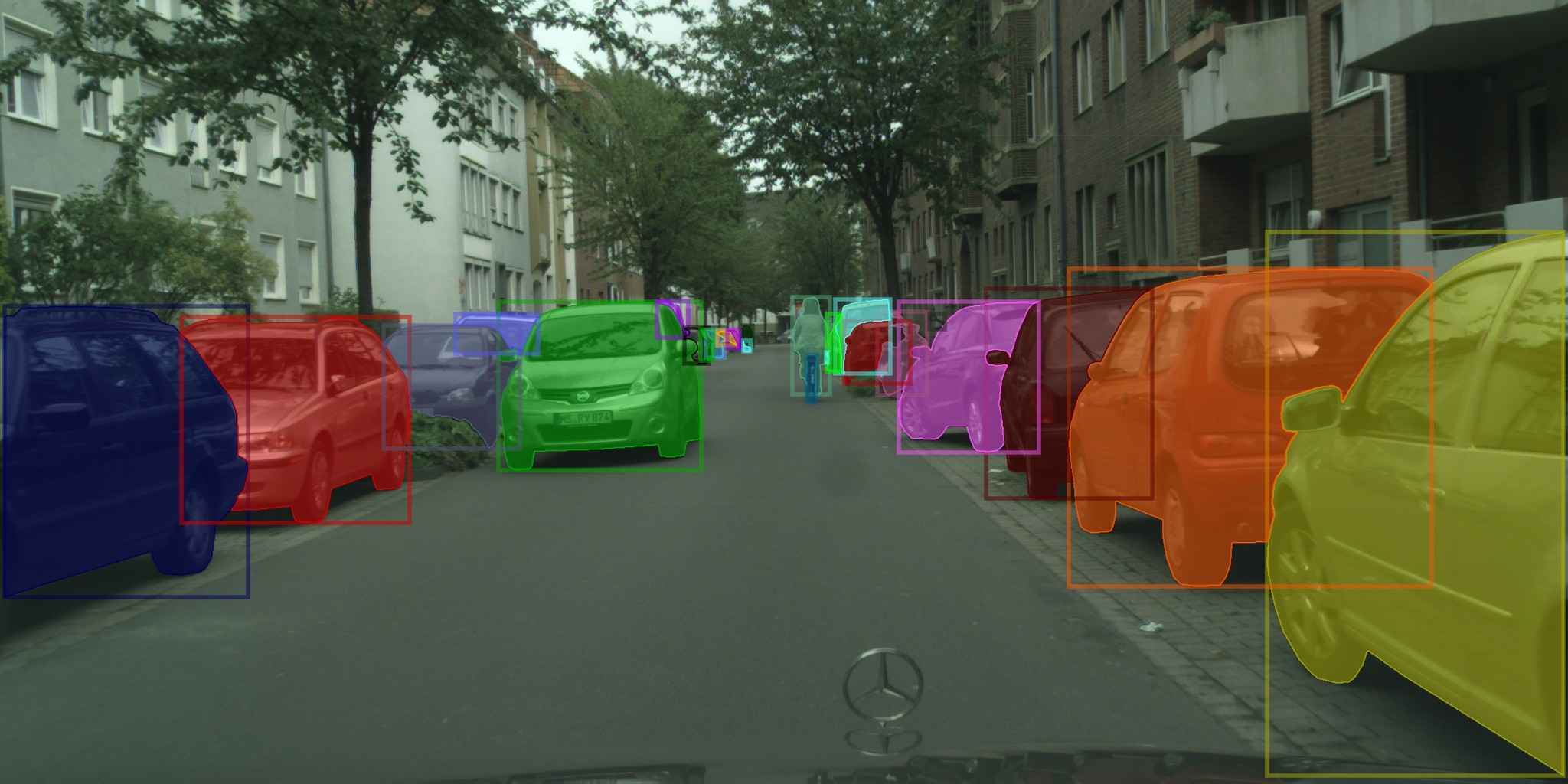}
		\label{fig:instseg:gt}
		\caption{GT}
	\end{subfigure}
	\begin{subfigure}[t]{0.22\textwidth}
		\centering
		\includegraphics[trim=0 0 0 100, clip=true, width=\linewidth]{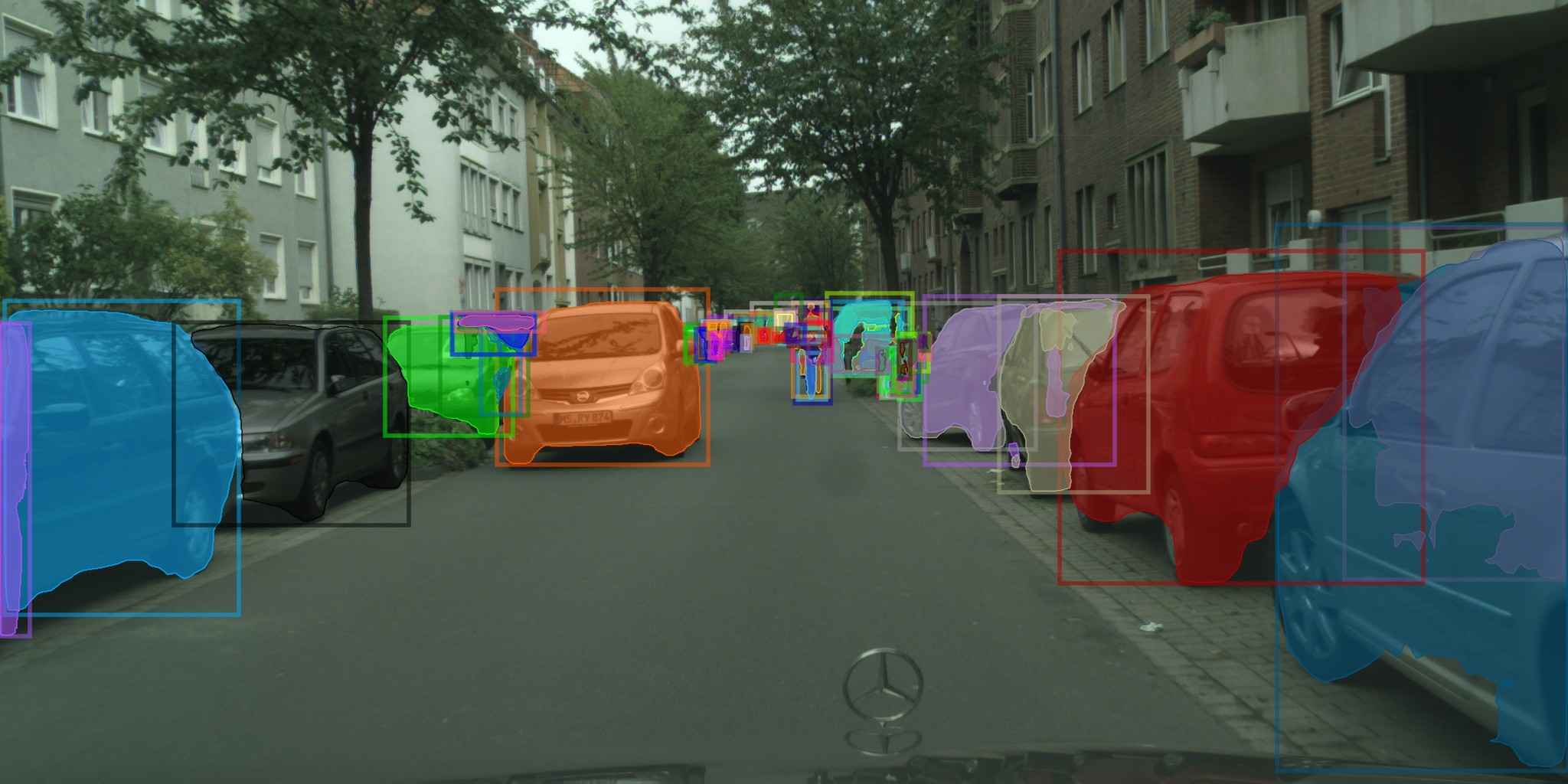}
		\label{fig:instseg:source_only}
		\caption{Source only}
	\end{subfigure}
	\begin{subfigure}[t]{0.22\textwidth}
		\centering
		\includegraphics[trim=0 0 0 100, clip=true, width=\linewidth]{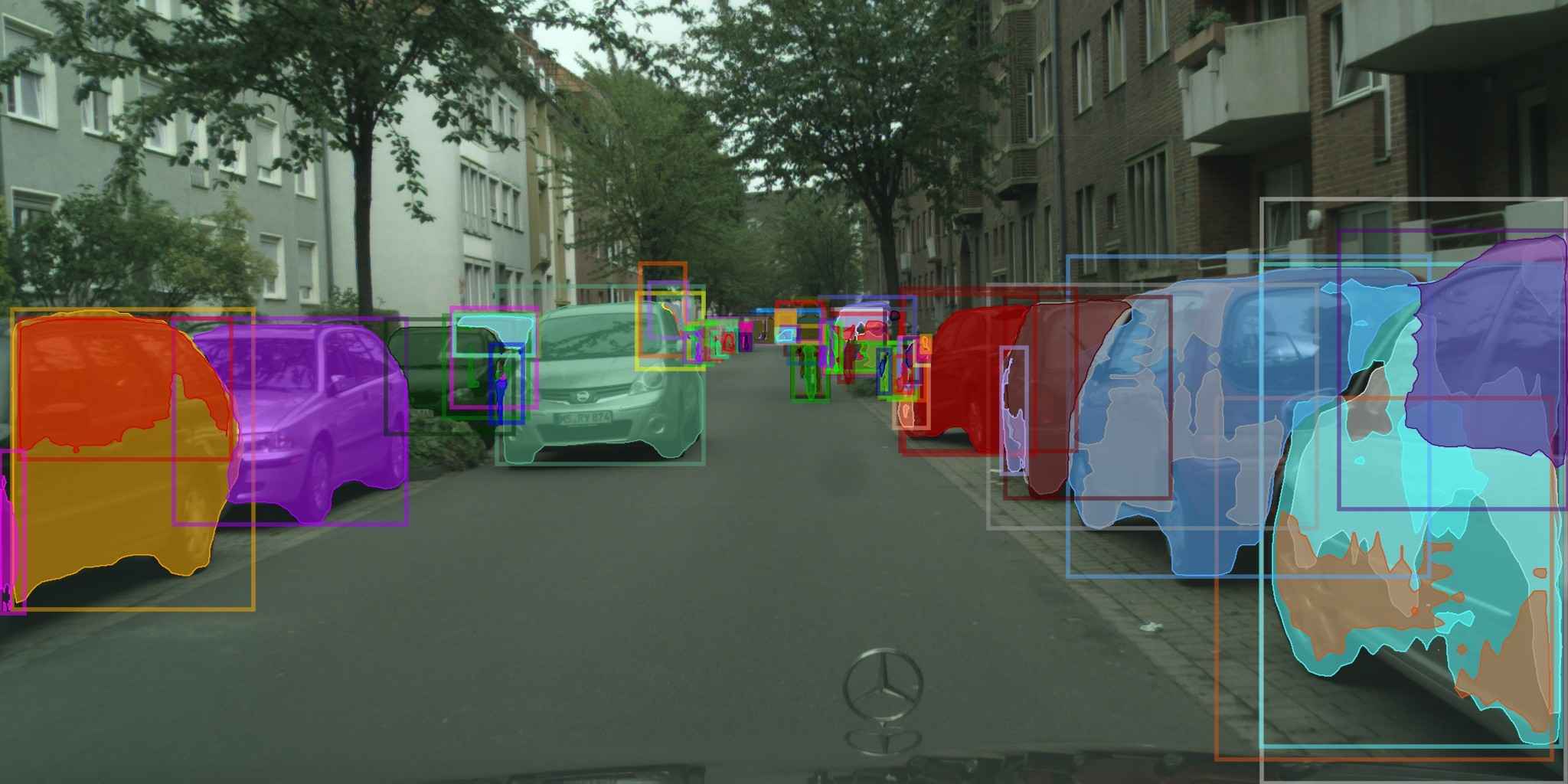}
		\label{fig:instseg:dafrcnn}
		\caption{DAFRCNN \cite{Chen2018CVPRd}}
	\end{subfigure}
	\begin{subfigure}[t]{0.22\textwidth}
		\centering
		\includegraphics[trim=0 0 0 100, clip=true, width=\linewidth]{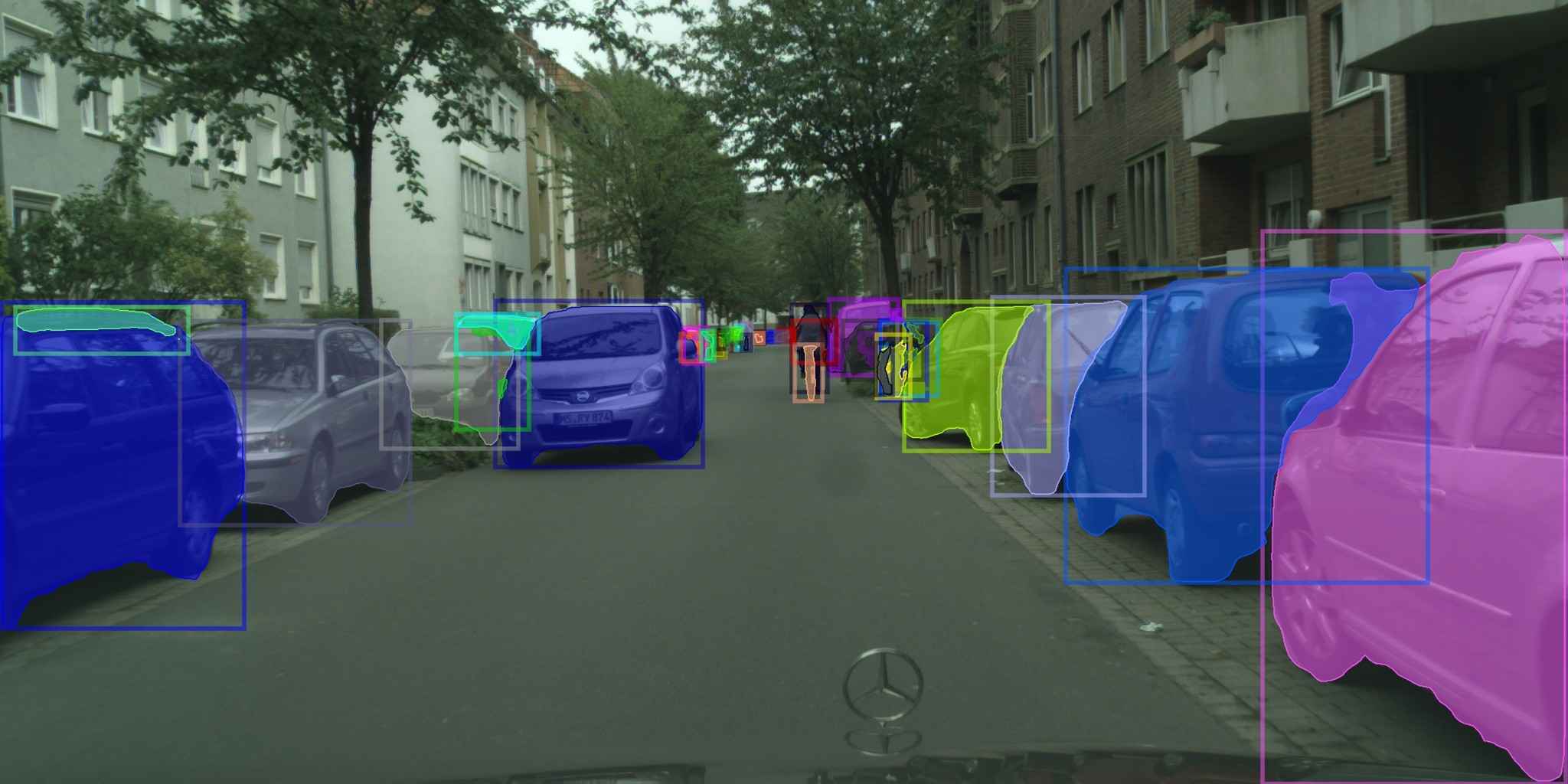}
		\label{fig:instseg:ours}
		\caption{WSJT + CWFA (Ours)}
	\end{subfigure}
	\caption{\textbf{Instance segmentation results (SYN $\rightarrow$ CS).} In contrast to our approach,
		both baselines exhibit a higher false positive rate. Besides improved detection performance, also the contours of our
		segmentations are typically more accurate.}
	\label{fig:instseg_comparison}
\end{figure*}

%% file: figures/box3d_comparison.tex
\begin{figure*}[tbp]
	\centering
	\begin{subfigure}[t]{0.22\textwidth}
		\centering
		\includegraphics[trim=0 0 0 100, clip=true,width=\linewidth]{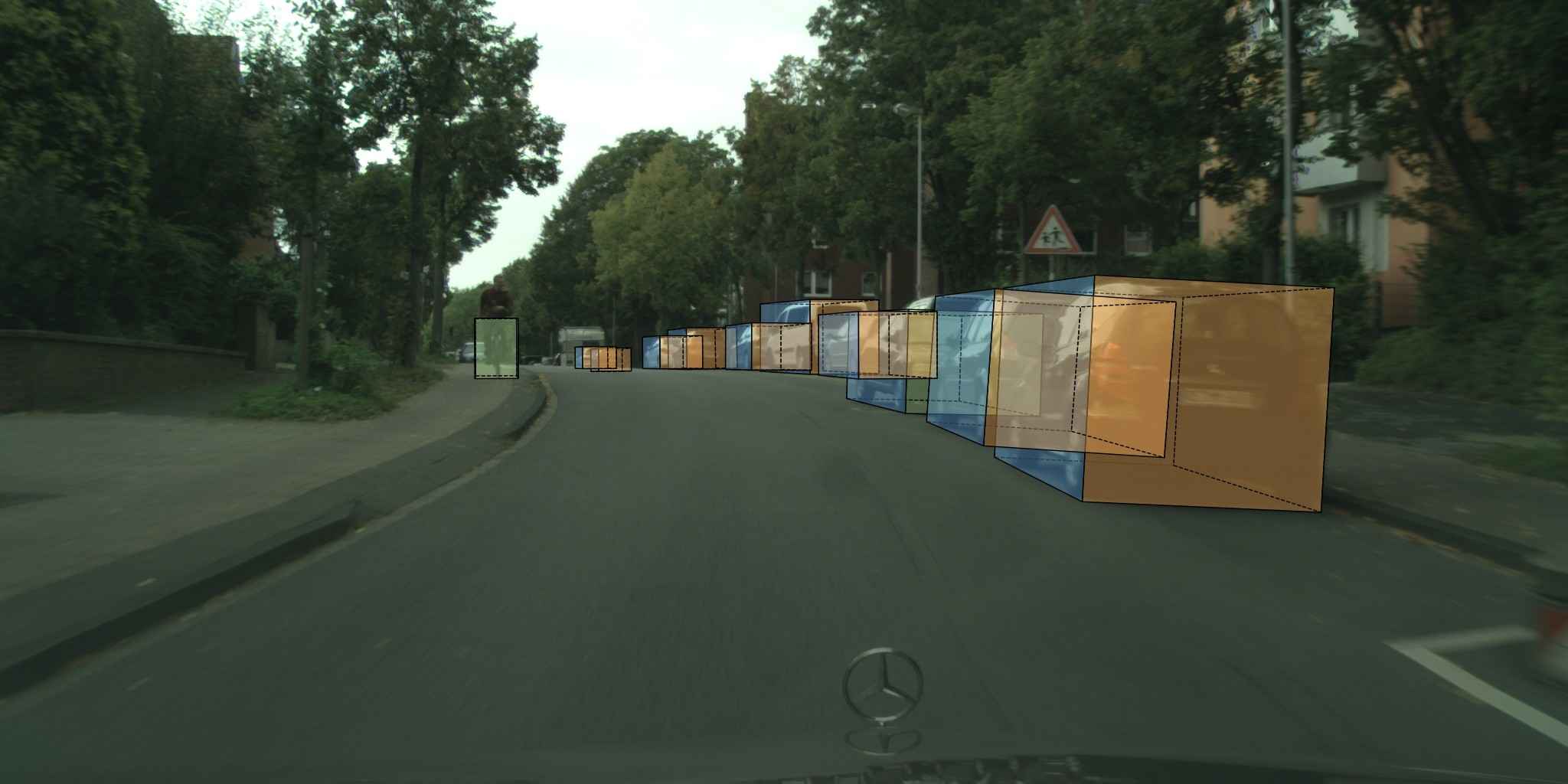}
	\end{subfigure}
	\begin{subfigure}[t]{0.22\textwidth}
		\centering
		\includegraphics[trim=0 0 0 100, clip=true,width=\linewidth]{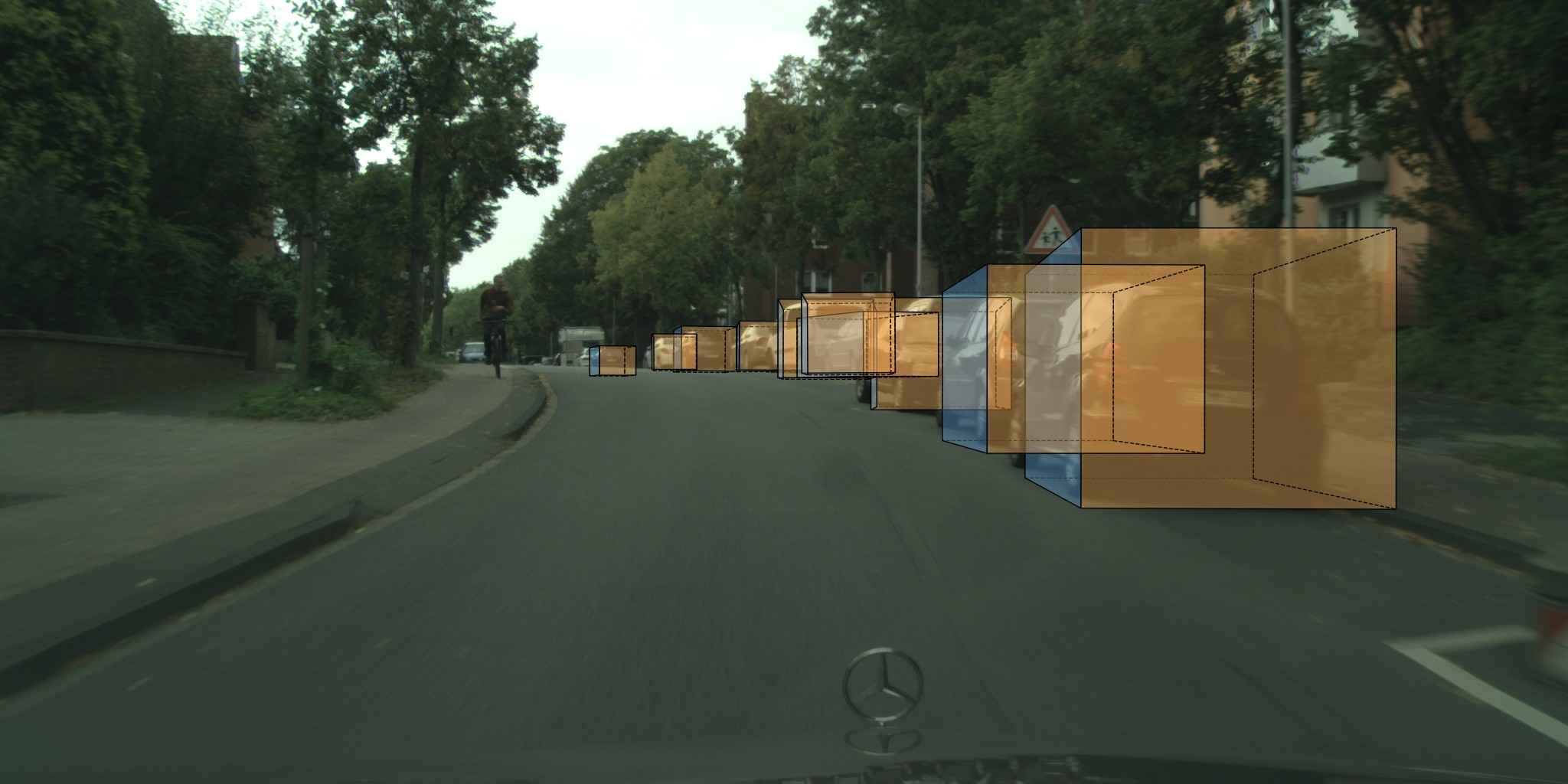}
	\end{subfigure}
	\begin{subfigure}[t]{0.22\textwidth}
		\centering
		\includegraphics[trim=0 0 0 100, clip=true,width=\linewidth]{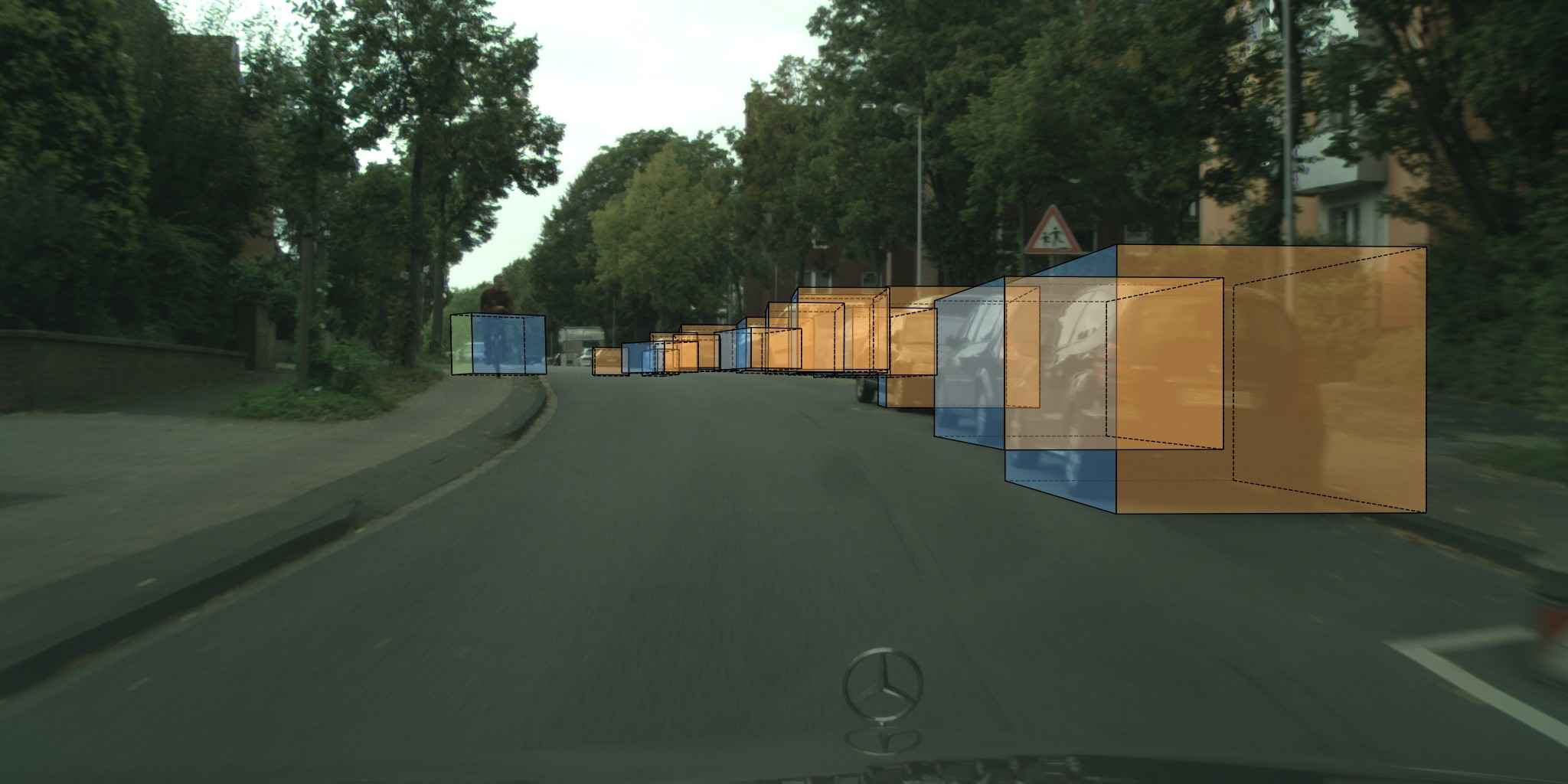}
	\end{subfigure}
	\begin{subfigure}[t]{0.22\textwidth}
		\centering
		\includegraphics[trim=0 0 0 100, clip=true,width=\linewidth]{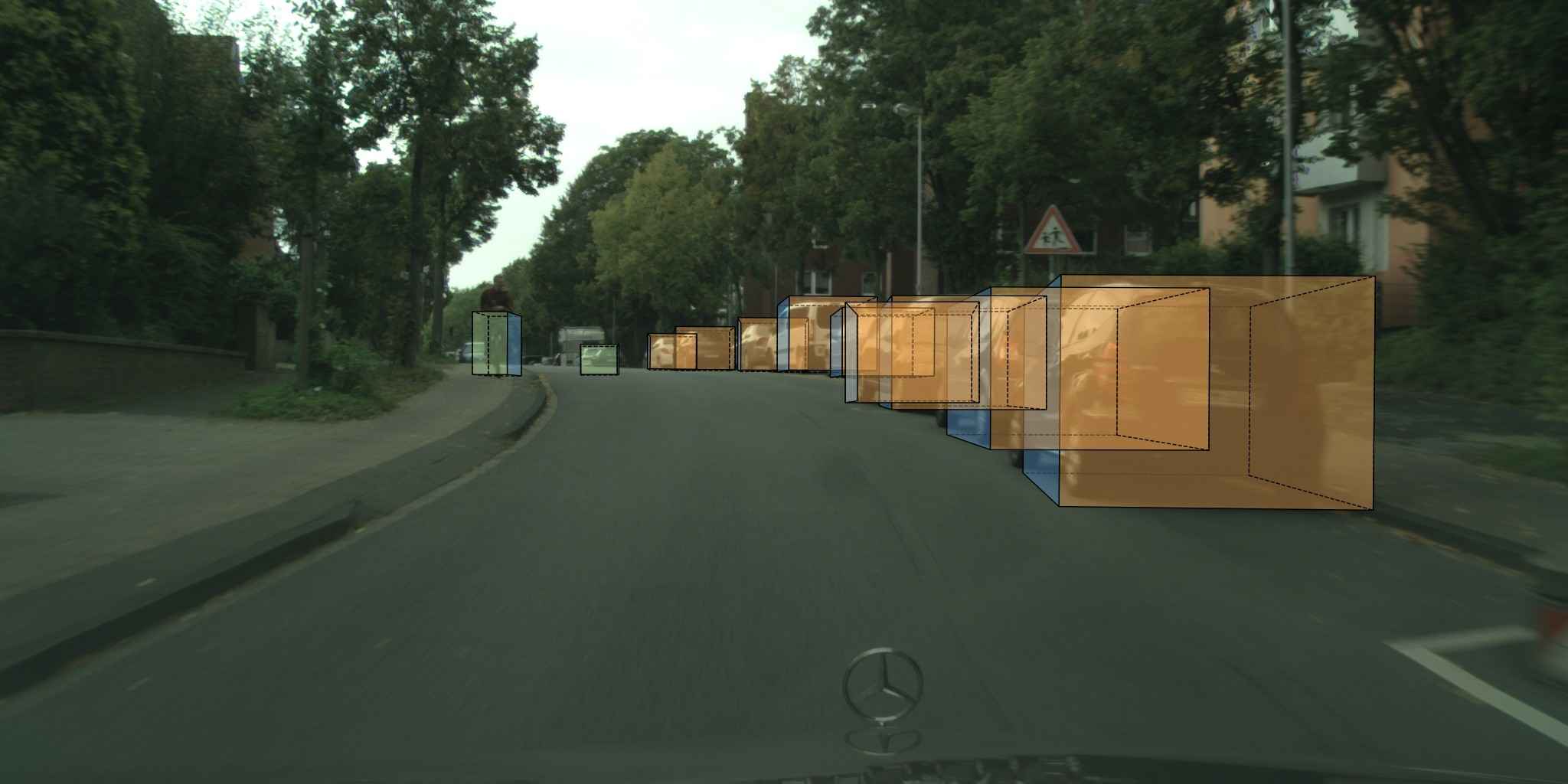}
	\end{subfigure}
	\begin{subfigure}[t]{0.22\textwidth}
		\centering
		\includegraphics[trim=800 200 0 300, clip=true, width=\linewidth]{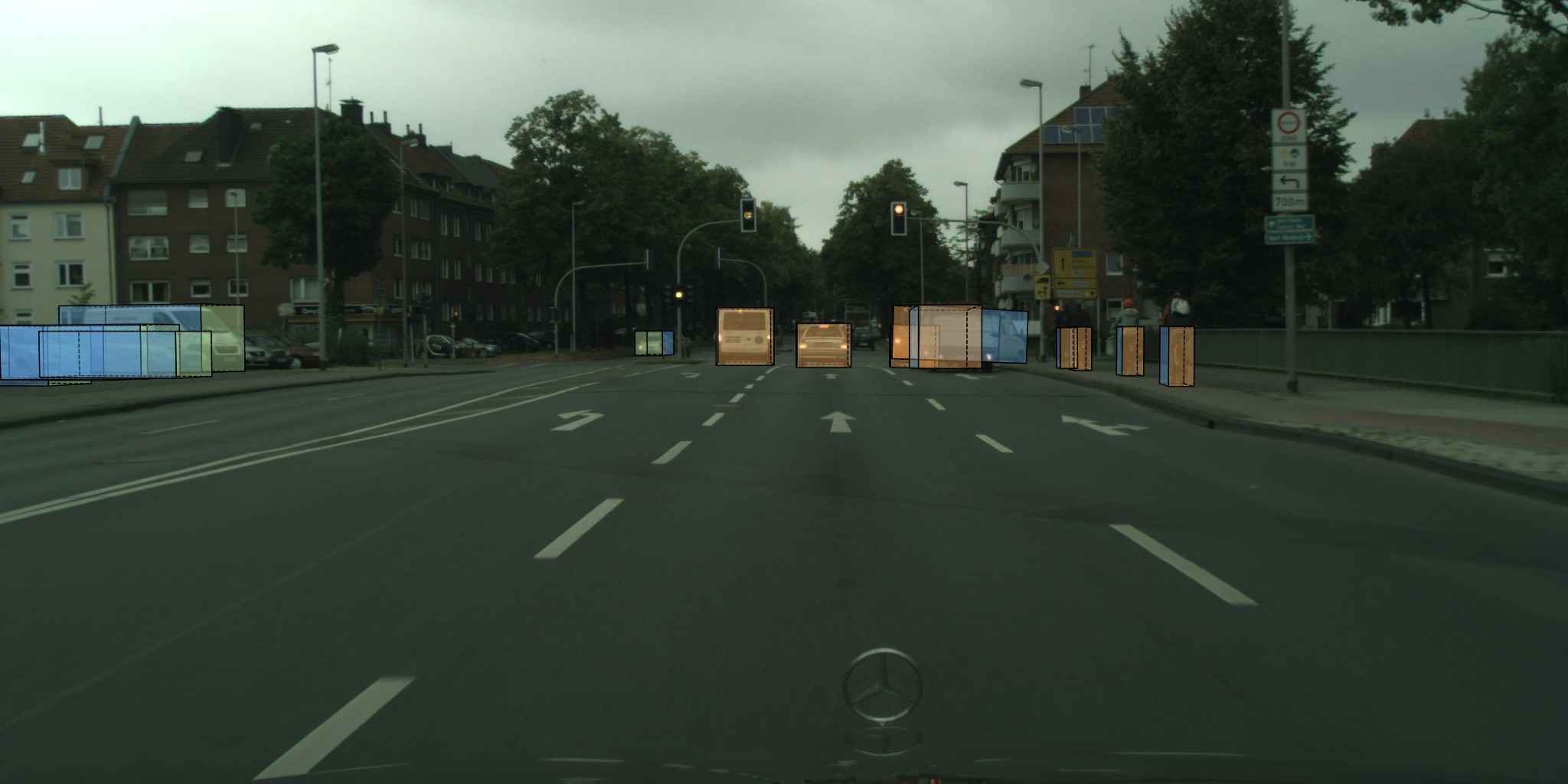}
	\end{subfigure}
	\begin{subfigure}[t]{0.22\textwidth}
		\centering
		\includegraphics[trim=800 200 0 300, clip=true, width=\linewidth]{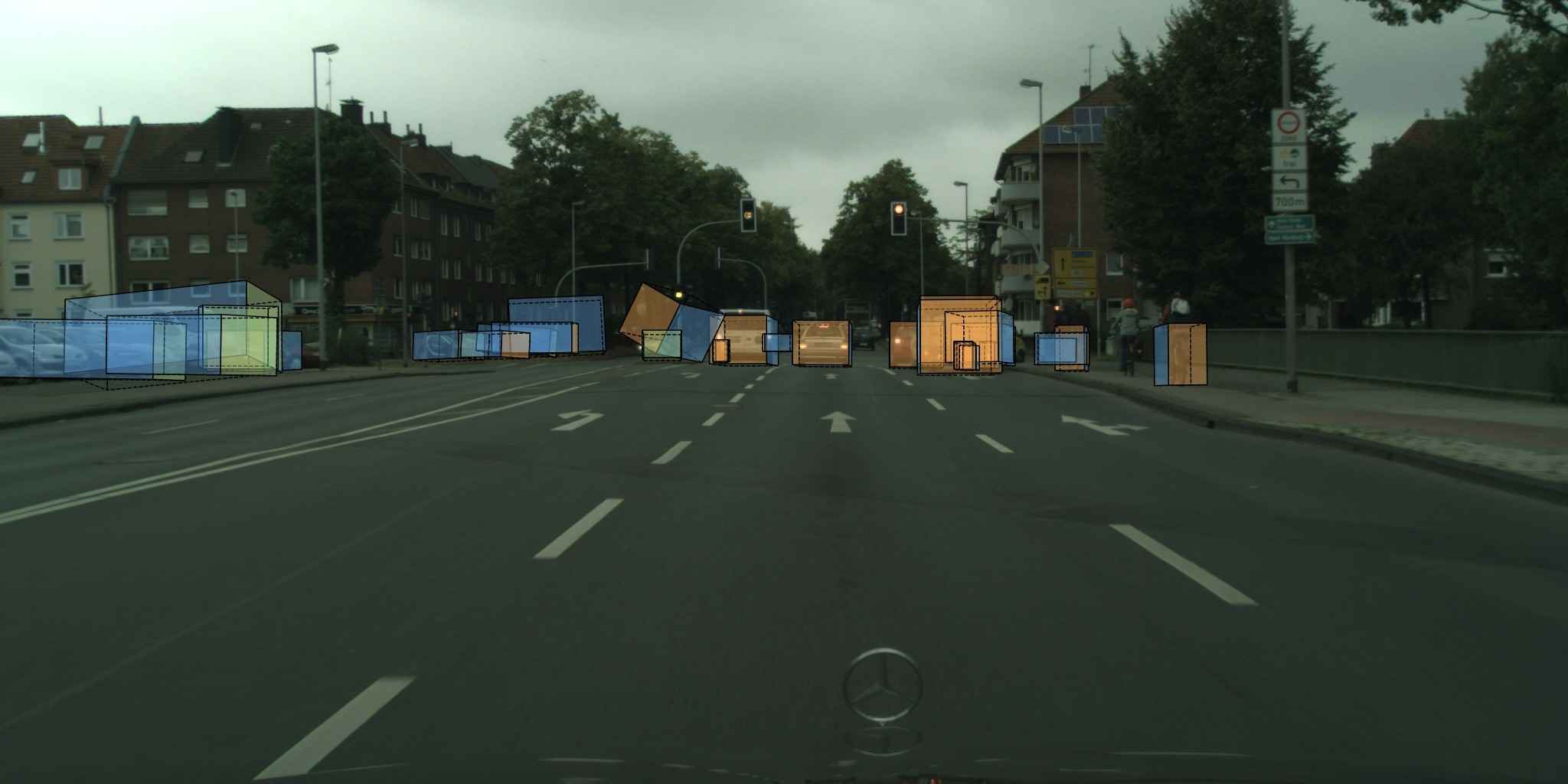}
	\end{subfigure}
	\begin{subfigure}[t]{0.22\textwidth}
		\centering
		\includegraphics[trim=800 200 0 300, clip=true, width=\linewidth]{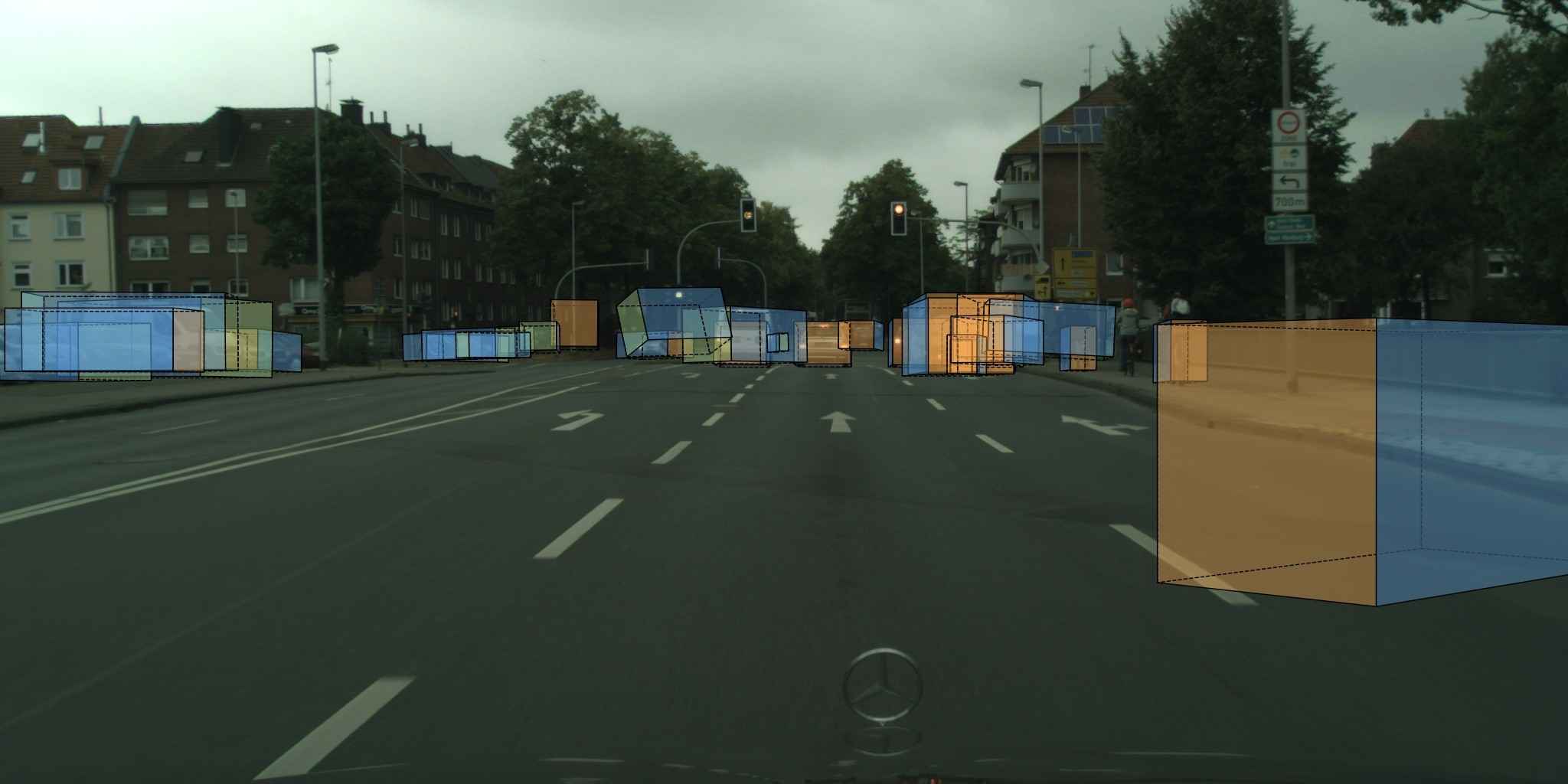}
	\end{subfigure}
	\begin{subfigure}[t]{0.22\textwidth}
		\centering
		\includegraphics[trim=800 200 0 300, clip=true, width=\linewidth]{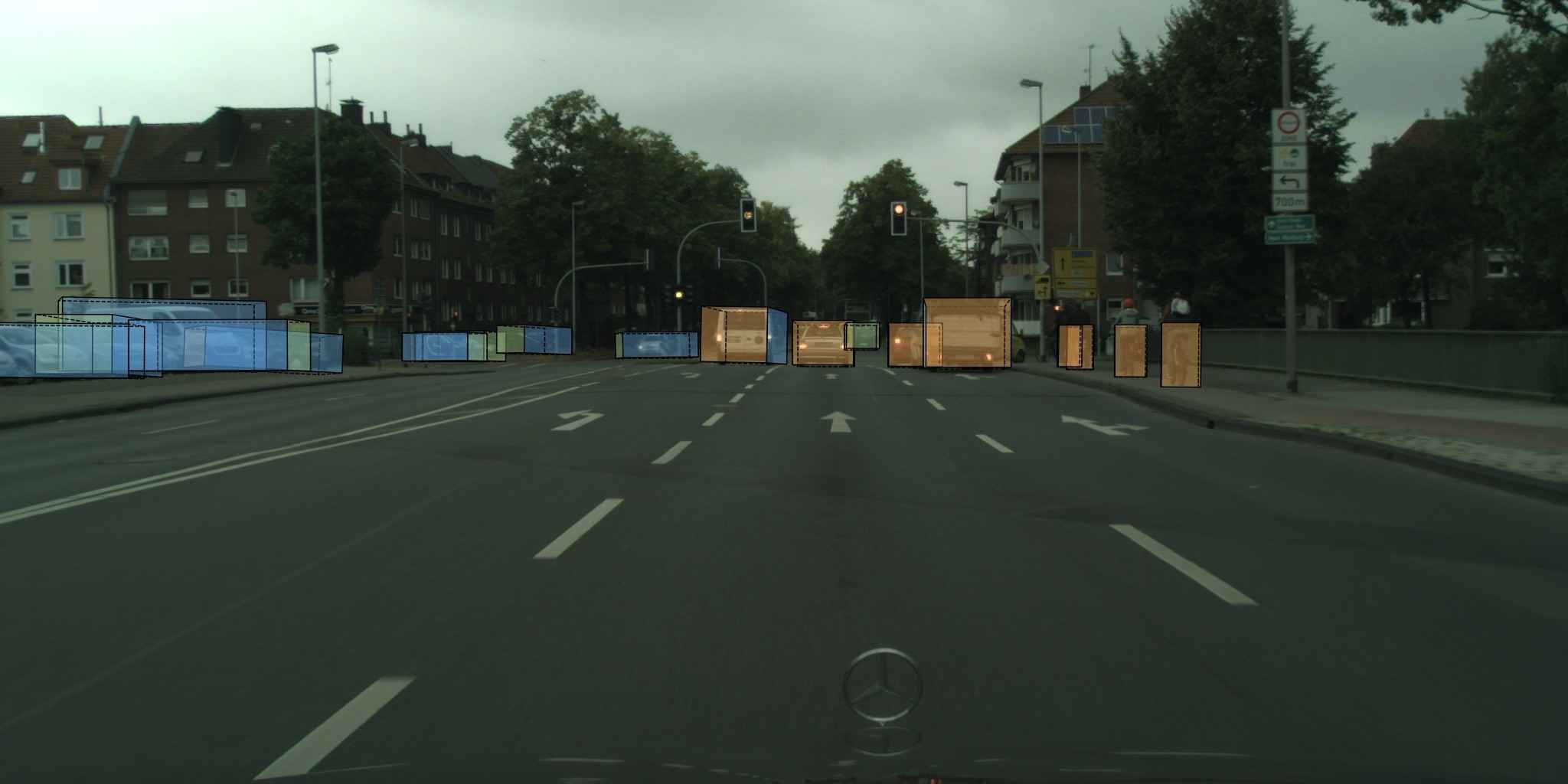}
	\end{subfigure}
	\begin{subfigure}[t]{0.22\textwidth}
		\centering
		\includegraphics[trim=0 0 0 100, clip=true,width=\linewidth]{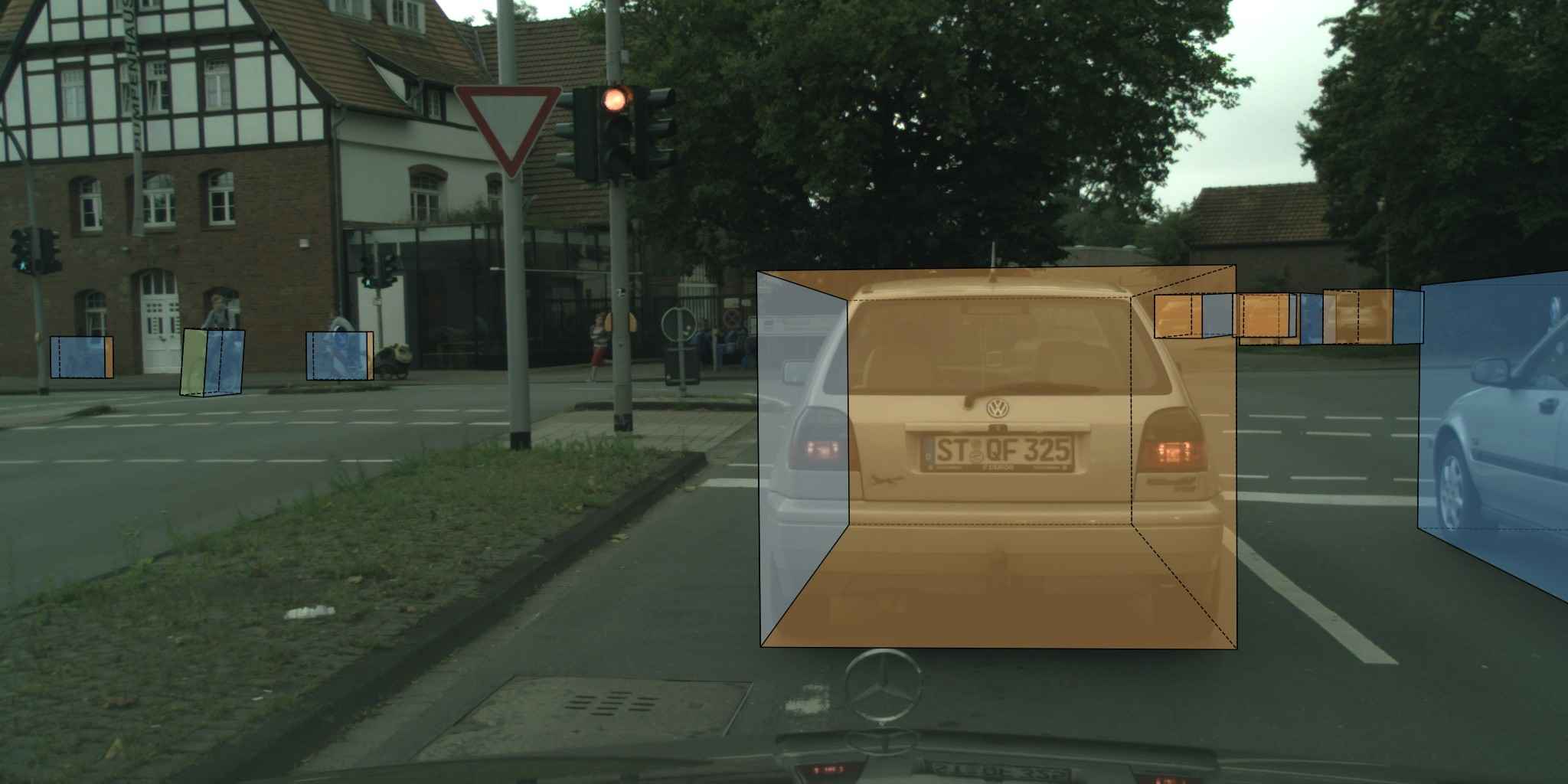}
		\label{fig:box3d:gt}
		\caption{GT}
	\end{subfigure}
	\begin{subfigure}[t]{0.22\textwidth}
		\centering
		\includegraphics[trim=0 0 0 100, clip=true,width=\linewidth]{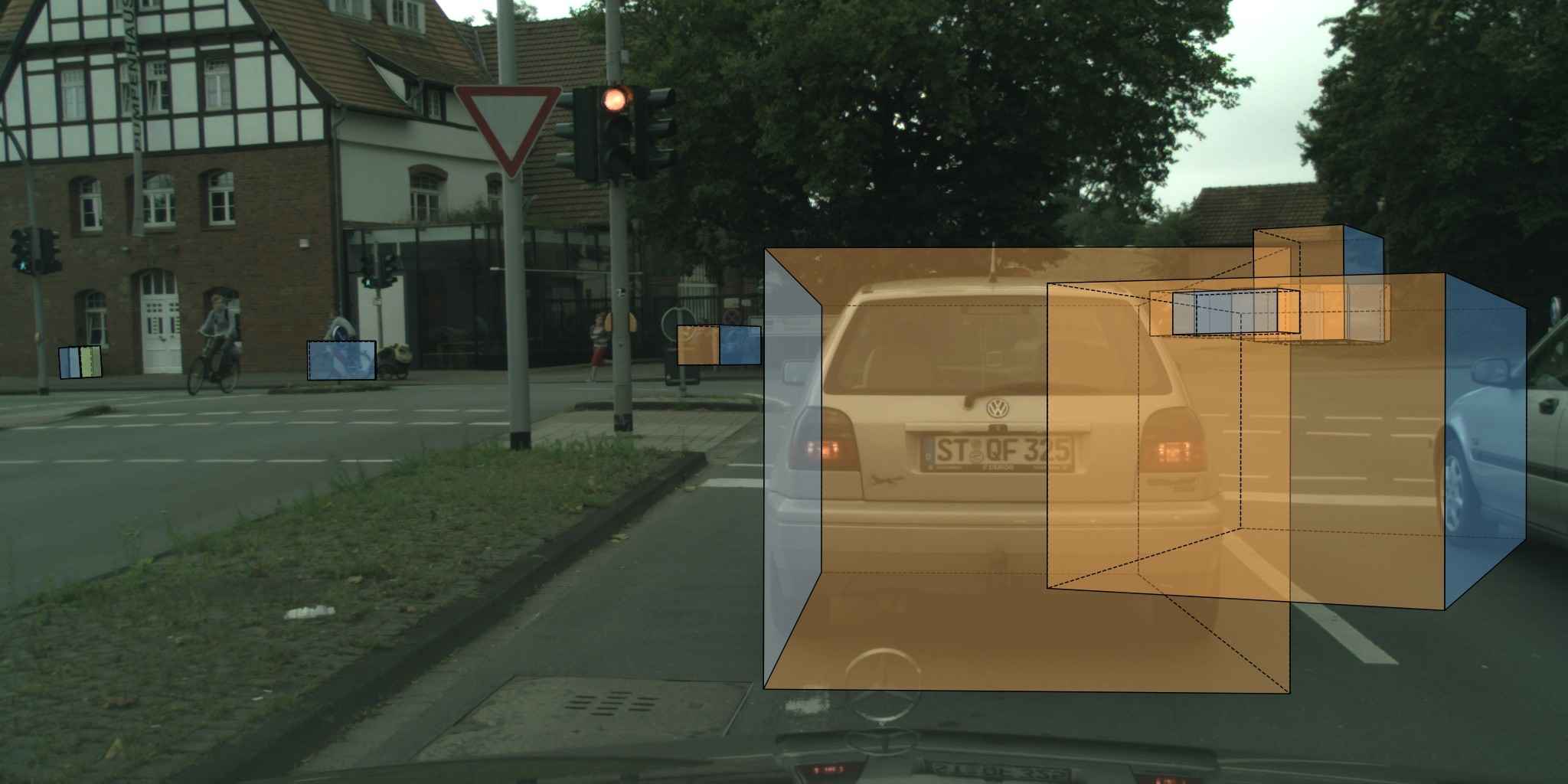}
		\label{fig:box3d:source_only}
		\caption{Source only}
	\end{subfigure}
	\begin{subfigure}[t]{0.22\textwidth}
		\centering
		\includegraphics[trim=0 0 0 100, clip=true,width=\linewidth]{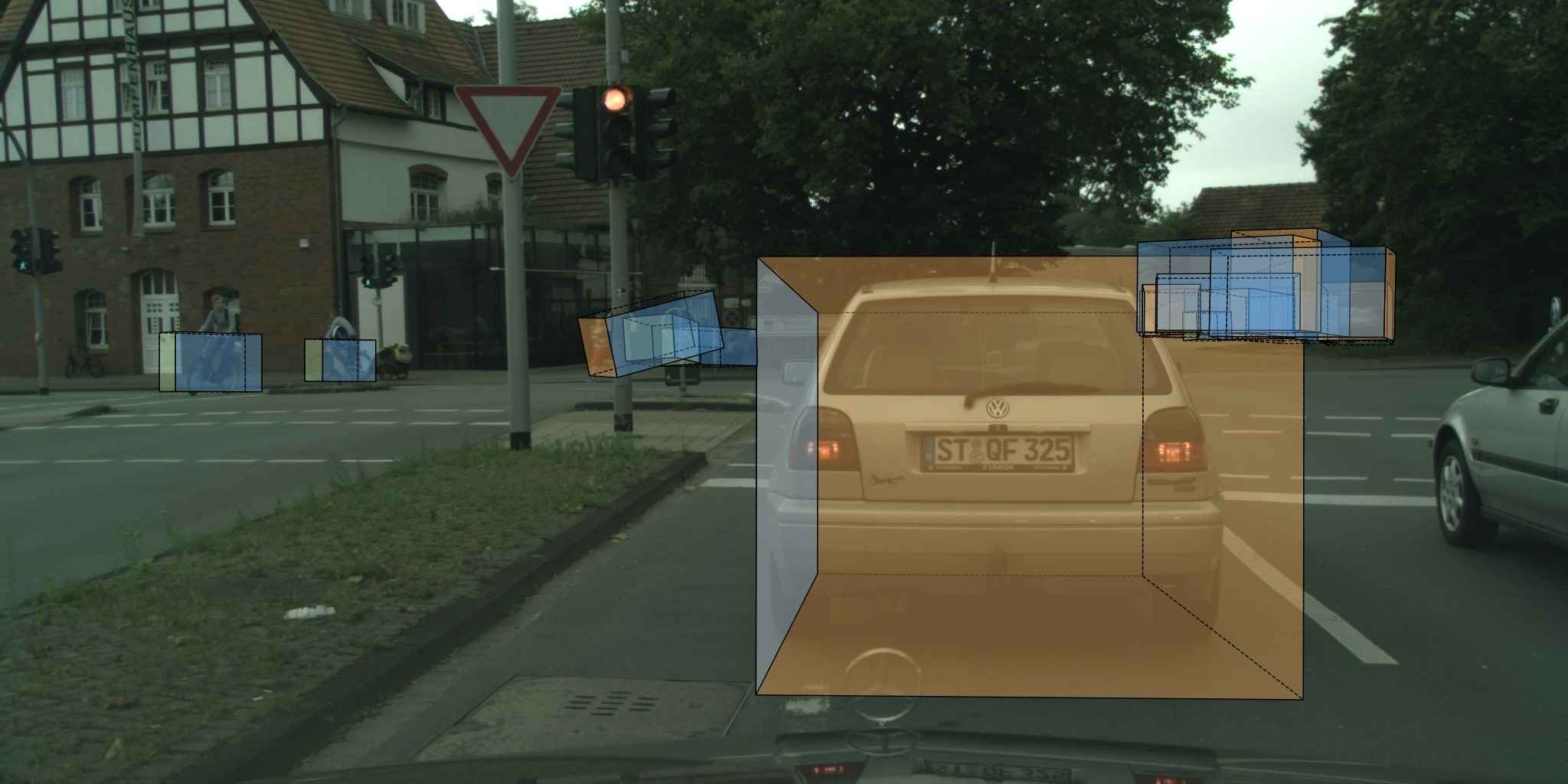}
		\label{fig:box3d:dafrcnn}
		\caption{DAFRCNN \cite{Chen2018CVPRd}}
	\end{subfigure}
	\begin{subfigure}[t]{0.22\textwidth}
		\centering
		\includegraphics[trim=0 0 0 100, clip=true,width=\linewidth]{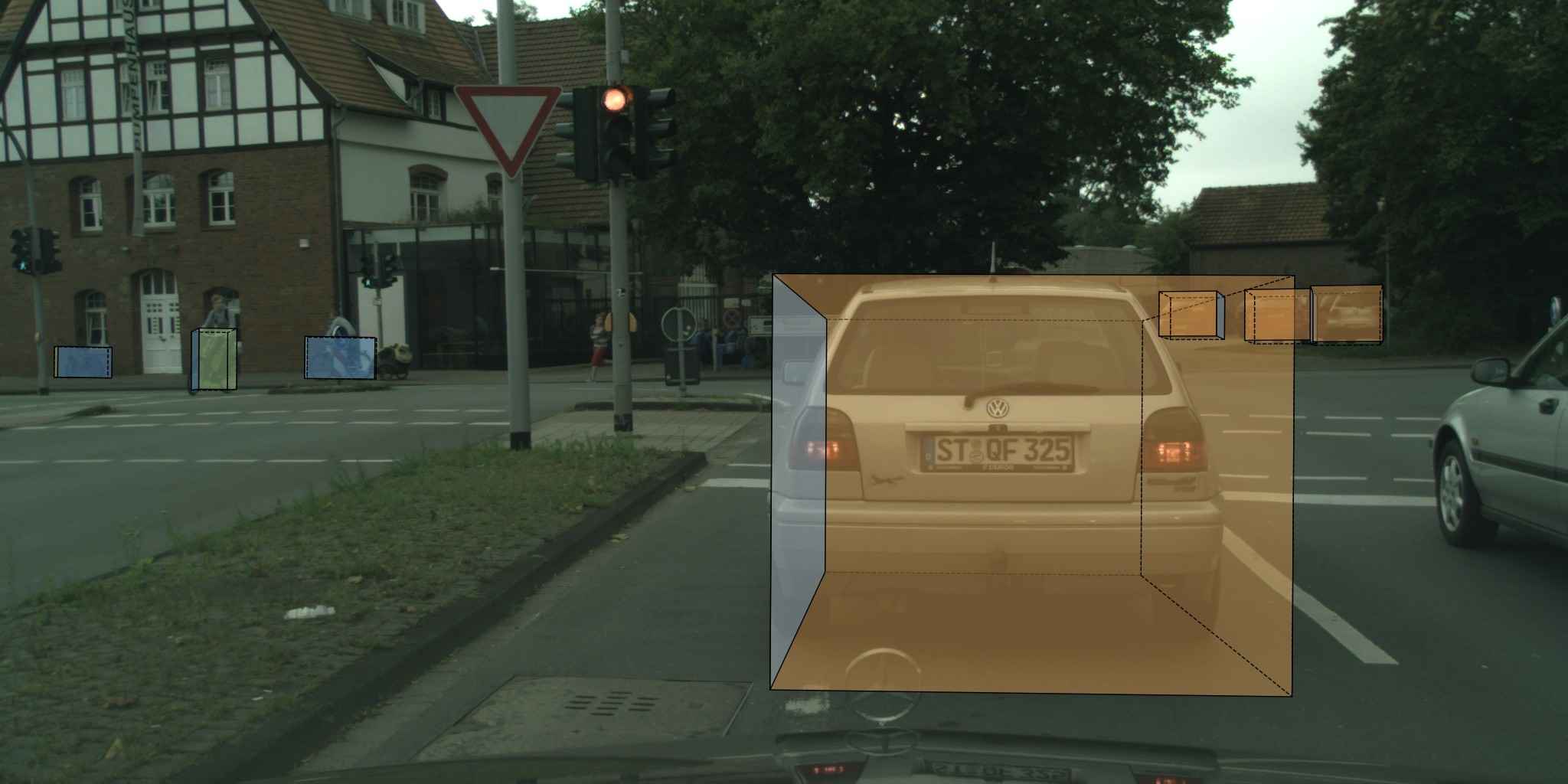}
		\label{fig:box3d:ours}
		\caption{WSJT + CWFA (Ours)}
	\end{subfigure}
	\caption{\textbf{Monocular 3D detection results (SYN $\rightarrow$ CS).} Besides less false positive detections,
	our approach also yields favorable orientation estimates in comparison to the baseline methods.}
	\label{fig:box3d}
\end{figure*}

%% file: results/instance_segmentation/all_datasets_all_methods.tex
\begin{table}[tbp]
	\footnotesize
	\centering

	\begin{tabular}{l c c c c c}
		\toprule
		\multirow{2}{*}{Method}     & \multicolumn{2}{c}{Target Supervision}  & SYN $\rightarrow$ & CS$\rightarrow$ & VIP $\rightarrow$ \\
		                            & 2D         & Mask                         & CS                & FCS             & CS                 \\
		\midrule
		Source Only                 &            &            & 17.0   & 13.9 & 6.9 \\
		DAFRCNN \cite{Chen2018CVPRd}&            &            & 17.5   & 23.3 & 9.1 \\
		SWDA \cite{Saito2019CVPR}   &            &            & 17.5   & 22.1 & 6.7 \\
		\midrule
		WSJT (Ours)                        & \checkmark &            & \textbf{32.1}   & \textbf{30.3} & \textbf{30.3}\\
		+ CWFA                      & \checkmark &            & 31.3   & \textbf{30.3} & 30.2\\
		\midrule
		Oracle                      & \checkmark & \checkmark & 33.6   & 30.2 & 36.1\\
		\bottomrule
	\end{tabular}

    \caption{\textbf{Results on instance segmentation.} We report the mean average precision.}
    \label{tab:instance:all}
\end{table}

%% file: results/bounding_box_3d/all_datasets_all_methods.tex
\begin{table}[tbp]
	\footnotesize
	\centering

	\begin{tabular}{l c c c c c}
		\toprule
		\multirow{2}{*}{Method}     & \multicolumn{2}{c}{Target Supervision}  & SYN $\rightarrow$              & CS $\rightarrow$                \\
		                            & 2D         & 3D                         & CS  & FCS \\
		\midrule
		Source Only                 &            &                                       & 14.2                                  & 13.7 \\
		DAFRCNN \cite{Chen2018CVPRd}&            &                            & 15.4                                  & 19.5 \\
		SWDA \cite{Saito2019CVPR}   &            &                            & 14.8                                  & 17.8 \\
		\midrule
		WSJT (Ours)                        & \checkmark &                            & 20.4                                  & 22.6 \\
		+ CAFA                      & \checkmark &                            & 21.1                                  & 24.1 \\
		+ CWFA                      & \checkmark &                            & \textbf{23.4}                                  & \textbf{24.2} \\
		\midrule
		Oracle                      & \checkmark & \checkmark                 & 25.0                                  & 24.7 \\
		\bottomrule
	\end{tabular}
	\caption{\textbf{Results on monocular 3D detection.} Values are mean Detection Scores \cite{Gaehlert2020ARXIV}.}
	\label{tab:detection_3d:all}
\end{table}

%% file: results/ablation_studies/box2d_gt_ablation_instance_segmentation.tex
\begin{table}[tbp]
	\footnotesize
	\centering

	\begin{tabular}{l c c}
		\toprule
		\multirow{2}{*}{Method}             & Instance Segmentation & 3D Detection\\
								            & (mAP)      & (mDS) \\
		\midrule
		Source Only                         & 27.7          & 22.8  \\
		DARCNN \cite{Chen2018CVPRd}         & 27.4          & 23.7 \\
		SWDA \cite{Saito2019CVPR}           & 27.2          & 24.5 \\
		\midrule
		WSJT (Ours)                                & \bf{36.2}     & 23.9 \\
		+ CWFA                         & 34.1          & \bf{31.6} \\
		\midrule
		Oracle                              & 45.5          & 34.5 \\
		\bottomrule
	\end{tabular}

	\caption{\textbf{Ablation Study.} To evaluate the transferability of the learned feature
	representations, we decouple the final performance from that of the 2D detection network, by
	using ground-truth 2D bounding boxes at test-time (SYN $\rightarrow$ CS).}

    \label{tab:ablation:2d_gt}
\end{table}

%% file: figures/failure_cases.tex
\begin{figure*}[tbp]
    \centering
    \begin{subfigure}[t]{0.22\textwidth}
      \centering
      \includegraphics[width=\linewidth]{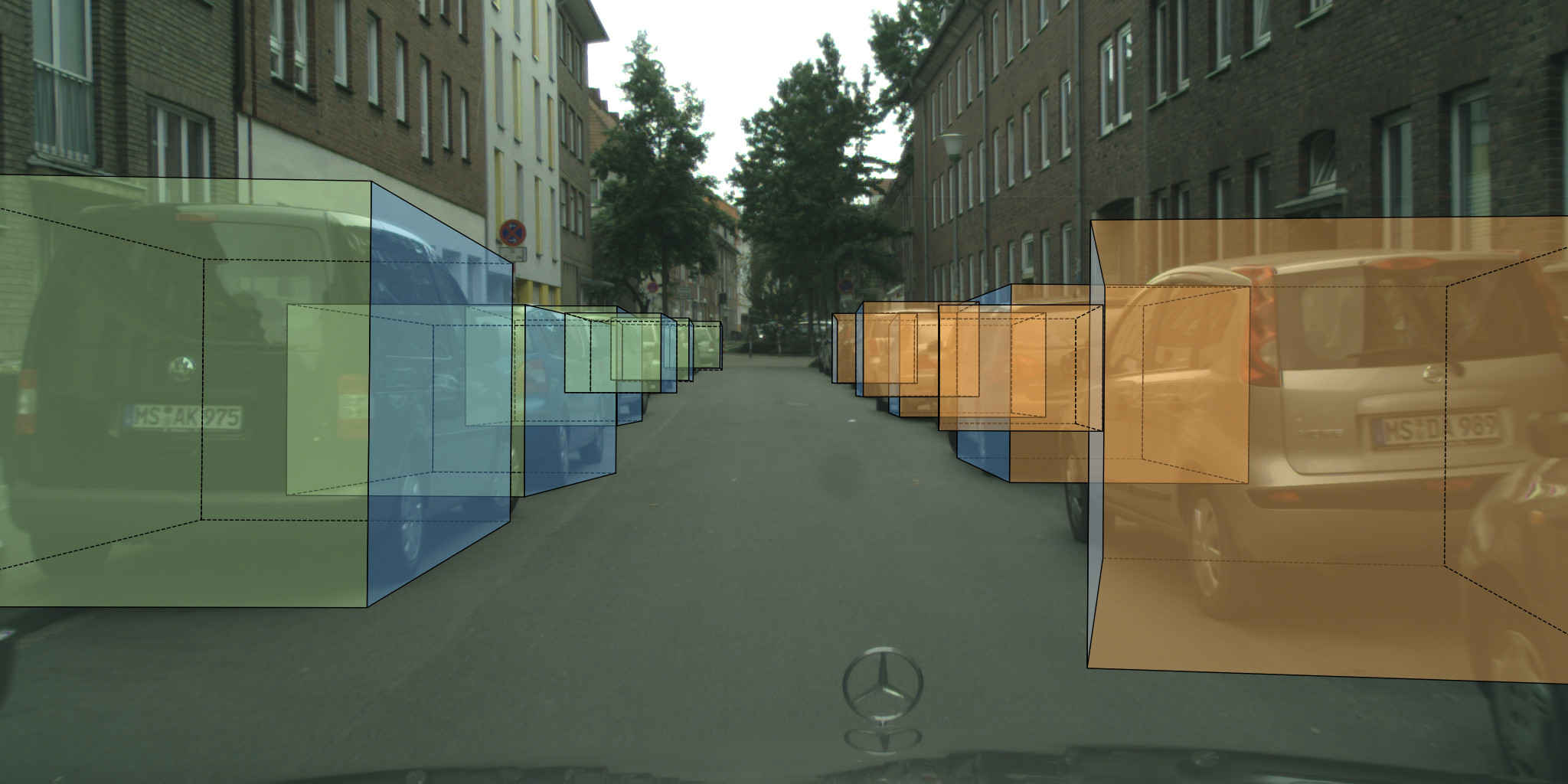}
      \label{fig:failure_case:oneway}
      \subcaption{}
    \end{subfigure}
    \begin{subfigure}[t]{0.22\textwidth}
      \centering
      \includegraphics[width=\linewidth]{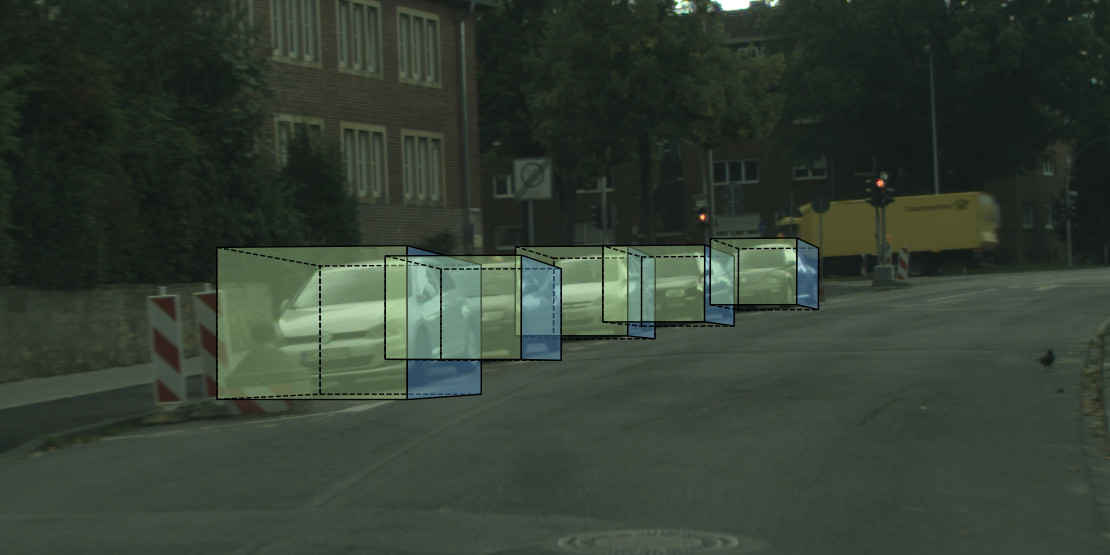}
      \label{fig:failure_case:slope}
      \subcaption{}
    \end{subfigure}
	\begin{subfigure}[t]{0.22\textwidth}
	\centering
	\includegraphics[width=\linewidth]{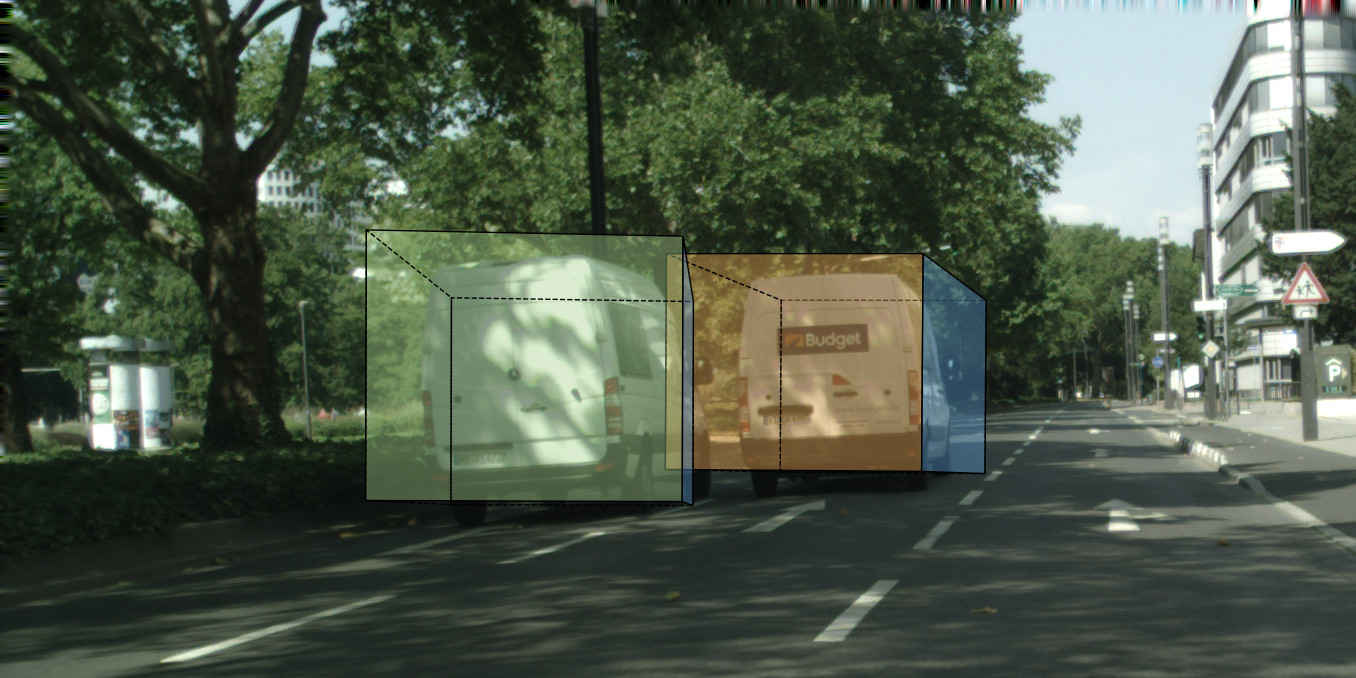}
	\label{fig:failure_case:orientation}
	\subcaption{}
	\end{subfigure}
	\begin{subfigure}[t]{0.22\textwidth}
	\centering
	\includegraphics[width=\linewidth]{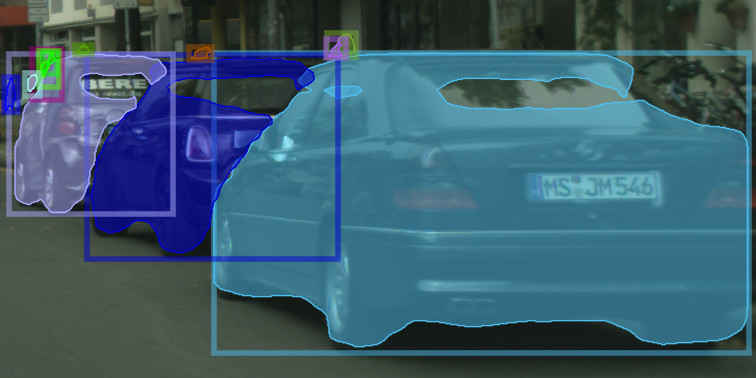}
	\label{fig:failure_case:window}
	\subcaption{}
	\end{subfigure}
    \caption{\textbf{Failure cases.}  We show failure cases for 3D detection (SYN $\to$ CS)
    and instance segmentation (VIP $\to$ CS) in \textit{(a)-(c)} and \textit{(d)}, respectively. \textit{(a)}: Due to the
    absence of one-ways in Synscapes, the orientations of all vehicles on the left hand side are flipped.
    \textit{(b)}: The lack of steep roads in the source domain results in a pitch bias.
    \textit{(c)}: Some vehicle types, \eg vans, are underrepresented in the source domain, causing spurious predictions.
    \textit{(d)}: Contrary to Cityscapes, vehicle windows are not included in the segmentation masks in VIPER, leading
    to false negatives.}
    \label{fig:failure_cases}
\end{figure*}

%% file: results/ablation_studies/map_over_time.tex
\begin{figure}[tbp]
	\centering
\begin{tikzpicture}
\begin{axis}[
    title={mAP over label time},
    xlabel={label time [h]},
    width=8.3cm, height=4.6cm,
    ylabel={mAP},
    xmin=0, xmax=500,
    ymin=15, ymax=35,
    xtick={0,100,200,300,400,500},
    ytick={15,20,25,30,35},
    legend pos=south east,
    ymajorgrids=true,
    grid style=dashed,
    line width=1pt,
    cycle list name=exotic,
    legend style={nodes={scale=0.6, transform shape}, at={(0.55,0.4)},anchor=west}
]

\definecolor{paper_blue}{rgb}{0.36, 0.6, 0.83}
\definecolor{paper_green}{rgb}{0.66, 0.82, 0.55}
\definecolor{paper_orange}{rgb}{0.96, 0.66, 0.33}

\addplot[mark=*, color=paper_orange]
coordinates {
	(0,17)(500,17)
};
\addplot[mark=*, color=paper_blue]
coordinates {
    (0,17.5)(500,17.5)
};

\addplot[mark=*, color=red]
coordinates {
	(8,23)(17,23)(80,30.5)(170,31.1)(500,32.1)
};

\addplot[mark=*, color=paper_green]
    coordinates {
    	(8,18)(17,20.1)(80,24.8)(170,28.7)(500,33.8)
    };

\legend{Source Only, Unsupervised \cite{Chen2018CVPRd, Saito2019CVPR}, WSJT + CWFA (Ours), Full Supervision}

\end{axis}
\end{tikzpicture}
\caption{\textbf{Ablation Study.} Instance segmentation performance (mAP) over label time (SYN $\rightarrow$ CS).}
\label{fig:map_over_time}
\end{figure}
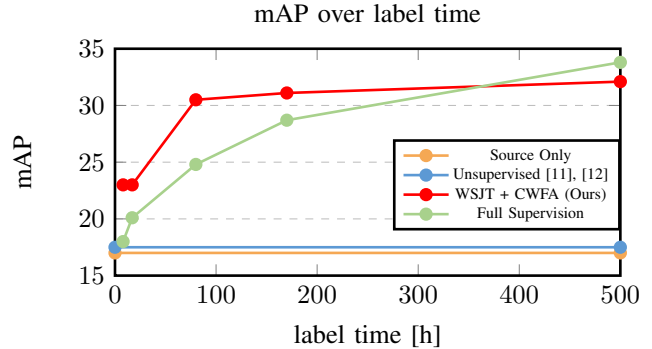

%% file: sections/conclusion.tex
\section{Conclusion}
In this work, we have presented a novel weakly-supervised domain adaptation setting, which exploits
the structure of cascaded detection tasks. In our experiments, we have demonstrated that models
adapted in our setting outperform unsupervised adaptation approaches by a large margin and can be
competitive with their fully supervised counterparts.
Although we considered instance segmentation
and monocular 3D detection as examples, we are convinced this setting can be applied to other
cascaded detection tasks such as human pose estimation, object tracking or trajectory forecasting.